\documentclass[preprint,12pt]{elsarticle}

\usepackage{amssymb}
\usepackage{amsmath}
\usepackage{amsfonts}
\usepackage{graphicx}
\usepackage{booktabs}
\usepackage{xcolor}
\usepackage{tcolorbox}
\usepackage{enumitem}
\usepackage[ruled,vlined]{algorithm2e}
\usepackage{float}              
\usepackage{framed}
\usepackage[section]{placeins}  
\usepackage{hyperref}

\journal{Environmental Modelling and Software}

\begin{document}

\begin{frontmatter}

\title{HydroGEM: A Self-Supervised Zero Shot Hybrid TCN-Transformer Foundation Model for Continental Scale Streamflow Quality Control}

\author[uvm-cs,uvm-wri]{Ijaz Ul Haq\corref{cor1}}
\ead{ihaq@uvm.edu}
\cortext[cor1]{Corresponding author}
\author[uvm-cs]{Byung Suk Lee}
\ead{bslee@uvm.edu}
\author[uvm-geo,uvm-wri]{Julia N. Perdrial}
\ead{jperdria@uvm.edu}
\author[uvm-wri]{David Baude}
\ead{dbaude@uvm.edu}
\affiliation[uvm-cs]{organization={Department of Computer Science, University of Vermont},
            city={Burlington}, 
            state={VT},
            country={USA}}
\affiliation[uvm-wri]{organization={Water Resources Institute, University of Vermont},
            city={Burlington}, 
            state={VT},
            country={USA}}
\affiliation[uvm-geo]{organization={Department of Geography and Geosciences, University of Vermont},
            city={Burlington}, 
            state={VT},
            country={USA}}
\begin{abstract}
Advances in sensor networks have enabled real-time, high-resolution stream discharge monitoring from in-situ gauges, yet persistent sensor malfunctions and data quality issues limit their utility. Manual quality control by expert hydrologists creates a bottleneck that delays scientific and operational use of these observations and cannot scale with networks generating millions of measurements annually. To address this gap, we introduce HydroGEM (Hydrological Generalizable Encoder for Monitoring), a foundation model for continental-scale streamflow quality control designed to support human expertise rather than replace it. HydroGEM uses a two-stage training approach: self-supervised pretraining on 6.03 million clean sequences from 3,724 USGS stations learns general hydrological representations, followed by fine-tuning with synthetic anomalies for detection and reconstruction. A hybrid Temporal Convolutional Network–Transformer architecture (14.2M parameters) captures both local temporal patterns and long-range dependencies, while hierarchical normalization handles learning across six orders of magnitude in discharge. On held-out real observations from 799 stations with synthetic anomalies spanning 18 types grounded in USGS operational standards, HydroGEM achieves F1 = 0.792 for detection and 68.7\% reconstruction error reduction, outperforming the strongest baseline by 36.3\%. For cross-national validation on 100 Environment and Climate Change Canada stations, we adopt tolerant evaluation with a $\pm$24 hour buffer to accommodate weak labels derived from operational corrections recorded with daily granularity. HydroGEM achieves Tolerant F1 = 0.70, with stable precision across all buffer sizes and 90.1\% segment-level detection of anomaly events, demonstrating cross-national generalization. The model maintains consistent detection across correction magnitudes (1–100\%) and aligns with operational seasonal patterns, with peak flagging rates during winter ice-affected periods matching hydrologists' correction behavior. Architectural separation between simplified training anomalies and complex physical-space test anomalies confirms that performance reflects learned hydrometric principles rather than pattern memorization.
\end{abstract}


\begin{highlights}
\item Continental-scale foundation model trained on 3,724 USGS sites with 6.03 million sequences
\item Two-stage self-supervised pretraining with synthetic anomaly injection for detection
\item Achieves F1 = 0.792 detection accuracy and 68.7\% reconstruction-error reduction
\item Zero-shot transfer to 100 Canadian ECCC stations achieves Tolerant F1 = 0.70
\item Human-in-the-loop design for operational quality control workflows
\end{highlights}


\end{frontmatter}
\section{Introduction}
\label{sec:introduction}

Real-time hydrological monitoring networks provide essential observations for water resources management, flood forecasting, ecosystem protection, and climate change adaptation~\cite{kratzert2019ungauged,kratzert2022caravan,sterle2024camelschem}. The United States Geological Survey operates more than 10,000 stream gauging stations that collectively produce millions of paired discharge and stage measurements each month, forming the backbone of decision-making across federal, state, and local agencies~\cite{kratzert2019towards,sterle2024camelschem}. High-frequency sensor data are only as useful as their quality, and without timely, accurate quality control, the growing volume of observations becomes a bottleneck rather than a resource~\cite{nearing2022lstm}.

Advances in sensor technology, telemetry, and network infrastructure have dramatically expanded the spatial and temporal density of hydrological monitoring. Modern gauging stations transmit stage and discharge at sub-hourly intervals, enabling real-time flood warnings, reservoir operations, and ecological flow management. Yet despite these technological improvements, raw sensor data remain inherently imperfect. Even in well-maintained monitoring networks, physical sensors inevitably experience drift, fouling, ice effects, rating-curve shifts, clock errors, and transmission failures, producing intermittent gaps or anomalies in the data~\cite{sun2019review,dogo2019survey}. These issues are an inherent part of sensing in operational environments, where the central challenge is ensuring fast, reliable quality control of the data they generate.

Traditional quality control in hydrological monitoring combines manual expert review with rule-based automated checks~\cite{aquarius2024,schmidt2023saqc,Jones2022}. Commercial systems such as AQUARIUS implement configurable range checks, rate-of-change thresholds, and statistical comparison filters, providing standardized workflows and audit trails~\cite{aquarius2024}. Open-source frameworks like SaQC advance reproducibility through sequential test configurations and traceable quality flags~\cite{schmidt2023saqc}. However, rule-based approaches face fundamental limitations: fixed thresholds require site-specific calibration and capture only obvious outliers. Context-dependent anomalies often evade detection, including ice effects that preserve plausible values but bias discharge~\cite{mcmillan2018uncertainty}, gradual sensor drift within historical ranges~\cite{taormina2015neural}, and rating-curve shifts after channel-altering floods~\cite{addor2017large}. Multivariate relationships between discharge and stage, critical for identifying coupled sensor failures or unit conversion errors, are difficult to encode in rule-based systems. As monitoring networks expand and temporal resolution increases, manual calibration becomes prohibitively expensive.

Current operational practice therefore relies heavily on domain experts who visually inspect time series, interpret stage–discharge relationships, and apply site-specific judgment~\cite{nearing2022lstm,goldstein2015peeking}. While this expert-driven workflow produces high-quality records, its manual nature cannot keep pace with expanding monitoring networks and rising data frequency. The resulting bottleneck stems not from insufficient expertise but from the growing mismatch between data volumes and the capacity for timely human review.

Recent advances in machine learning offer a promising path toward more efficient and scalable hydrologic data quality control. Modern models can process continuous data streams, capture subtle temporal dependencies, and operate consistently across large station networks. These capabilities align well with the needs of contemporary monitoring systems. In principle, such methods could provide rapid, reliable quality screening across thousands of sites.

Classical anomaly detection methods have been applied to environmental monitoring with mixed success. Statistical approaches (z-score tests, ARIMA residuals), distance-based methods (k-nearest neighbors, isolation forests), and one-class classifiers (one-class SVM, autoencoders) provide baseline capabilities~\cite{belay2023unsupervised,shi2023groundwater}. Recent applications include wavelet-based multi-resolution analysis for river stage~\cite{nourani2014applications} and comparative evaluations of isolation forests versus one-class SVM for groundwater monitoring, where OCSVM achieved 88\% precision on synthetic data~\cite{shi2023groundwater}.

Deep learning approaches show stronger performance by capturing complex temporal dependencies. LSTM networks and Transformer attention mechanisms excel at modeling long-range patterns in time series~\cite{hochreiter1997long,he2019temporal}. Temporal Convolutional Networks enable efficient parallel processing with exponentially large receptive fields~\cite{han2023twawdlstm}. Hybrid designs that combine convolutional temporal feature extraction with attention based global context have also been explored for time series forecasting and anomaly detection~\cite{wang2025tcntransformer,mulia2024kbjnet}. Autoencoders, including variational and LSTM-augmented variants, detect anomalies via reconstruction errors~\cite{osman2024variational,lee2024lstm}. Generative Adversarial Networks improve robustness through learned representations of normal data distributions~\cite{chen2025gasn}. Synthetic anomaly generation has emerged as a key technique for training when domain-specific labeled data are unavailable~\cite{kim2024enhancing,song2025synthetic}.

Translating this potential into practice, however, is challenging because continental-scale deployment faces two fundamental obstacles. First, hydrological systems are extremely heterogeneous: discharge spans six orders of magnitude across sites, and flow generation mechanisms vary with climate, physiography, and human modification~\cite{kratzert2019towards,kratzert2022caravan,lees2021hydrological}. Models typically train and evaluate on individual sites or small regional datasets (often fewer than 50 sites), failing to capture the extreme heterogeneity spanning snowmelt-dominated montane systems to ephemeral desert washes to regulated lowland rivers. Transfer learning across such diverse regimes remains largely unexplored in hydrological anomaly detection. Second, labeled anomaly datasets are scarce. Agency quality flags are not designed as ground truth labels that cleanly separate sensor faults from physical phenomena, and comprehensive labeled datasets across thousands of sites do not exist~\cite{dogo2019survey,belay2023unsupervised}.

Foundation models provide a path forward~\cite{devlin2019bert,dooley2023forecastpfn}. These large networks are pretrained on massive datasets with self-supervised objectives, then transferred to diverse downstream tasks. In weather and climate science, ClimaX demonstrated that transformers pretrained on heterogeneous climate reanalysis data with masked token prediction effectively transfer to forecasting, downscaling, and multi-variable prediction across spatiotemporal scales~\cite{nguyen2023climax}. Weather foundation models including FourCastNet~\cite{pathak2022fourcastnet}, GraphCast~\cite{lam2023graphcast}, and Pangu-Weather~\cite{bi2023accurate} now achieve medium-range forecast skill competitive with operational numerical weather prediction at orders of magnitude lower computational cost. Industrial-scale efforts such as IBM/NASA's Prithvi WxC (2.3B parameters)~\cite{prithvi2024wxc} and Oak Ridge's ORBIT ($>$100B parameters)~\cite{wang2024orbit} are extending this paradigm to multi-variable Earth system prediction.

In hydrology, Kratzert and colleagues pioneered multi-basin learning by training LSTM networks across hundreds of catchments using the CAMELS dataset~\cite{kratzert2019towards}, demonstrating that models learn generalizable runoff generation processes that transfer to ungauged basins. Subsequent work confirmed that diverse training across basins improves predictions despite regional hydrological variation~\cite{gauch2021proper}. Remote sensing foundation models pretrained on satellite imagery have shown strong transfer to environmental monitoring tasks~\cite{ali2025environmental}. Recent perspectives advocate for foundation models as assistive tools in hydrometeorology when developed with appropriate domain constraints~\cite{ali2025environmental}.

Yet foundation model principles have not been systematically applied to quality control for \emph{in situ} sensor networks. This gap reflects unique challenges: extreme scale heterogeneity (six orders of magnitude in discharge within a single dataset, requiring specialized normalization); irregular multi-site time series (non-gridded data with site-specific rating curves, instrument types, and operational protocols); real-time deployment constraints (inference must operate without access to ground truth); deploy-safe requirements (the model must not corrupt valid data, a failure mode unacceptable in operational settings); and human oversight mandates (agencies require interpretable outputs, uncertainty quantification, and audit trails for regulatory compliance). While hybrid convolution attention backbones are effective in general time series settings~\cite{wang2025tcntransformer,mulia2024kbjnet}, they typically do not address deploy-safe quality control requirements across thousands of heterogeneous gauging sites, including extreme magnitude variation, label scarcity, and cross-agency transfer constraints.

Our approach addresses these limitations by integrating advances from multiple domains. From foundation models~\cite{nguyen2023climax,pathak2022fourcastnet}, we adopt self-supervised pretraining on massive unlabeled datasets. From large-sample hydrology~\cite{kratzert2019towards,gauch2021proper}, we use multi-site learning for cross-basin transfer. From recent deep learning work~\cite{dooley2023forecastpfn,kim2024enhancing}, we use synthetic anomaly injection for task-specific fine-tuning. From operational QC systems~\cite{aquarius2024,schmidt2023saqc}, we incorporate human-in-the-loop workflows and audit requirements. The result is a system that learns multivariate patterns from continental-scale data, enabling context-aware detection without manual threshold tuning while preserving the human-in-the-loop workflows that agencies require.

To address these challenges, we introduce \textbf{HydroGEM} (Hydrological Generalizable Encoder for Monitoring), a self-supervised foundation model for continental-scale streamflow quality control. HydroGEM qualifies as a foundation model because it: (1) trains on massive diverse data (3,724 sites spanning six orders of magnitude in discharge), (2) uses self-supervised pretraining to learn general hydrological representations, (3) shows zero-shot transfer to unseen sites and countries, and (4) uses a modular architecture that allows task-specific adaptation.

HydroGEM uses a two-stage training approach. \textbf{Stage 1} pretrains a hybrid TCN-Transformer backbone on 6.03 million clean sequences from 3,724 USGS sites using masked reconstruction. Unlike many prior hybrid architectures that use conventional softmax attention~\cite{wang2025tcntransformer,mulia2024kbjnet}, our global module adopts cosine attention with a retention style temporal decay prior to improve stability on long sequences~\cite{mongaras2025cottention,sun2023retnet}. A hierarchical normalization scheme combines log transforms, site-specific standardization, and explicit scale embeddings to allow learning across extreme magnitude ranges while preserving physically meaningful scale structure~\cite{goodfellow2016deep}. \textbf{Stage 2} fine-tunes the pretrained backbone with a detection head using on-the-fly synthetic anomaly injection rather than curated labels~\cite{dooley2023forecastpfn,kim2024enhancing}. The training injector applies simplified corruptions (drift, spikes, flatlines, dropouts, clock shifts) in normalized space to encourage learning of fundamental hydrometric consistency principles rather than memorization of specific anomaly signatures. Clean data preservation exceeds 97\%, ensuring the model does not corrupt valid observations.

Evaluation emphasizes generalization across three dimensions. On held-out real observations from 799 USGS sites with synthetic anomalies spanning 18 types grounded in USGS operational standards, we create a four-axis separation (geographic, mathematical, temporal, parameter) that prevents pattern memorization. HydroGEM achieves F1 = 0.792 for anomaly detection and reduces reconstruction error by 68.7\% relative to injected corruptions, outperforming the strongest baseline by 36.3\%. Suggested corrections require expert review~\cite{Jones2022} before operational use. For cross-national validation, we evaluate zero-shot transfer to 100 Environment and Climate Change Canada (ECCC) sites using weak labels derived from operational corrections. Because correction records are often applied with daily granularity rather than precise anomaly boundaries, and correction methodologies vary across agencies~\cite{Jones2022}, we adopt tolerant evaluation with a $\pm$24 hour buffer as the primary metric for this dataset. HydroGEM achieves Tolerant F1 = 0.70, with stable precision across all buffer sizes from $\pm$1 to $\pm$24 hours. At the segment level, the model detects 90.1\% of anomaly events, demonstrating effective cross-national generalization despite differences in instrumentation and operational protocols.

HydroGEM is designed for human-in-the-loop workflows: a three-tier flagging system passes high-confidence clean data with minimal review, prioritizes uncertain observations for expert inspection, and provides suggested reconstructions with uncertainty estimates. All outputs include provenance information suitable for audit and feedback, supporting hydrologist expertise~\cite{Jones2022,nearing2022lstm,goldstein2015peeking}.

Minimal inference materials, runnable notebooks, and supporting documentation for the USGS synthetic benchmark and ECCC site set are available at \url{https://huggingface.co/Ejokhan/HydroGEM}.

In summary, this work advances hydrological quality control through five contributions:

\begin{enumerate}[leftmargin=*,itemsep=2pt]
    \item A continental-scale foundation model trained on 3,724 USGS sites with 6.03 million sequences, an order of magnitude larger than prior multi-site hydrological studies.
    
    \item A two-stage training approach combining self-supervised pretraining on clean data with synthetic anomaly injection for detection and reconstruction, reducing dependence on labeled anomalies.
    
    \item Hierarchical normalization enabling learning across six orders of magnitude while preserving scale-dependent physical behavior.
    
    \item Rigorous evaluation on held-out real observations from 799 sites with synthetic anomalies spanning 18 types grounded in USGS operational standards, achieving F1 = 0.792 for detection and 68.7\% reconstruction error reduction.
    
    \item Demonstrated cross-national transfer from USGS (USA) to ECCC (Canada) data with Tolerant F1 = 0.70, validating learned representations beyond the training distribution.
\end{enumerate}

Table~\ref{tab:positioning} contrasts HydroGEM with related approaches across key dimensions.

\begin{table}[htbp]
\centering
\small
\caption{Positioning HydroGEM relative to related approaches}
\label{tab:positioning}
\resizebox{\textwidth}{!}{%
\begin{tabular}{@{}lcccccc@{}}
\toprule
\textbf{Approach} & \textbf{Sites} & \textbf{Self-sup.} & \textbf{Zero-shot} & \textbf{Synthetic} & \textbf{Recon.} & \textbf{Deploy-safe} \\
\midrule
Rule-based QC~\cite{aquarius2024,schmidt2023saqc} & Any & N/A & N/A & No & Manual & Yes \\
Single-site ML~\cite{he2019temporal,lee2024lstm} & 1--10 & Varies & No & Sometimes & Varies & No \\
Multi-basin hydro~\cite{kratzert2019towards} & 100--500 & No & Yes & No & No & N/A \\
Weather FM~\cite{nguyen2023climax,pathak2022fourcastnet} & Gridded & Yes & Yes & No & No & N/A \\
\textbf{HydroGEM} & \textbf{3,724} & \textbf{Yes} & \textbf{Yes} & \textbf{Yes} & \textbf{Suggested} & \textbf{Yes} \\
\bottomrule
\end{tabular}%
}
\end{table}

Box 1 summarizes key technical terms used throughout this paper.

\noindent
\begin{minipage}{\textwidth}
\begin{framed}
\noindent\textbf{Box 1: Terminology}
\begin{itemize}[leftmargin=*,itemsep=1pt]
    \item \textbf{Clean data}: USGS approved records with quality code `A' (verified and approved by USGS hydrologists as accurate representations of observed conditions; used for Stage 1 training)
    \item \textbf{Masked data}: Clean data with random timesteps hidden for self-supervised reconstruction (Stage 1)
    \item \textbf{Corrupted data}: Clean data with synthetic anomalies injected (Stage 2 fine-tuning)
    \item \textbf{Anomalous data}: Real operational data requiring correction (deployment target)
    \item \textbf{Suggested reconstruction}: Model-proposed correction requiring hydrologist approval
    \item \textbf{Stage/Gage height}: Water surface elevation measured at the gauging station, typically in feet
    \item \textbf{Discharge}: Volume of water flowing past a point per unit time, typically in cubic feet per second (ft$^3$/s)
    \item \textbf{Rating curve}: Empirical relationship between stage and discharge at a specific site
\end{itemize}
\end{framed}
\end{minipage}


\section{Data and Evaluation Framework}
\label{sec:data}

\begin{figure}[htbp]
    \centering
    \includegraphics[width=\textwidth]{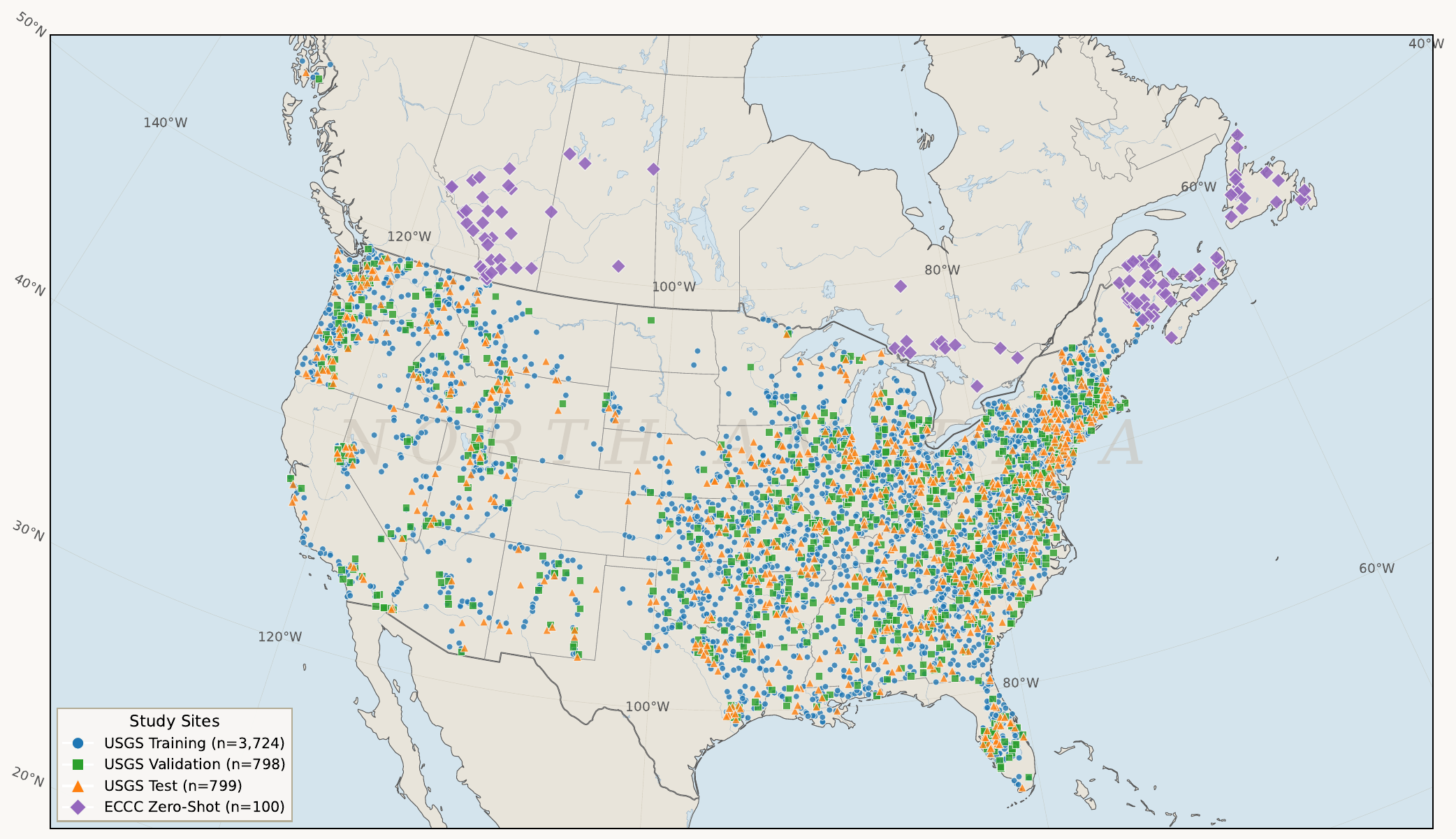}
    \caption{Geographic distribution of study sites. USGS stations are split into training (n = 3,724), validation (n = 798), and test (n = 799) sets with no geographic overlap. Canadian ECCC stations (n = 100) were selected based on data completeness and quality criteria described in Section~\ref{sec:canadian}.}
    \label{fig:data_map}
\end{figure}

Our evaluation strategy combines 3 data sources (Figure~\ref{fig:data_map}): (1) a continental-scale USGS corpus for foundation model pretraining, (2) synthetic test sets with anomalies grounded in USGS operational standards for controlled generalization assessment, and (3) real-world Canadian stations for zero-shot cross-national transfer validation. This design enables evaluation both under controlled conditions with known ground truth and under operational conditions that reflect agency quality control workflows.

\subsection{Foundation Training Corpus}
\label{subsec:training_corpus}

\subsubsection{USGS Streamflow Archive}
\label{subsubsec:usgs_archive}

We initially retrieved candidate records from 6,000 USGS streamgaging stations via the National Water Information System (NWIS)~\cite{usgs2024nwis}, covering January 2000 through December 2024. After applying the quality control and completeness filters described in Section~\ref{subsubsec:qc_preprocessing}, 679 stations were excluded, yielding 5,321 retained sites for analysis and partitioning. For each station, we selected the best 10-year continuous period with high-quality paired discharge ($Q$) and stage ($H$) observations. Best was defined as the continuous 10-year span maximizing paired $Q$ and $H$ completeness after quality control while minimizing gap filling. All retained records carry USGS qualification code `A' (approved), distinguishing them from provisional data subject to revision. Instantaneous readings (typically 15-minute intervals) were aggregated to hourly resolution via arithmetic means for consistent temporal scale.

The 10-year site-specific selection strategy balances data quality, temporal depth, and infrastructure consistency. Post-2000 focus ensures modern instrumentation across the network, while 10-year spans capture multiple seasonal cycles and extreme hydroclimatic events (droughts, floods) essential for robust anomaly detection. This period also corresponds to relatively standardized USGS quality control protocols, reducing confounding institutional factors.

\subsubsection{Spatial and Hydrologic Coverage}
\label{subsubsec:spatial_coverage}

The network spans the conterminous United States, Alaska, and Hawaii, providing true continental-scale geographic diversity (approximately 163$^\circ$W to 67$^\circ$W and 20$^\circ$N to 62$^\circ$N). Drainage areas range from 0.03 to 932,800 km$^2$ (median 196 km$^2$). Gauge elevations span $-69$ m to 2,747 m (mean 467 m), capturing coastal, lowland, montane, and alpine monitoring sites.

This spatial coverage ensures exposure to principal North American flow generation mechanisms: snowmelt-dominated montane regimes, rainfall-dominated humid systems, mixed snow-rain transitional climates, ephemeral arid flows, and regulated downstream conditions~\cite{chow1988applied}. Such hydrologic diversity is critical for developing transferable representations that generalize beyond specific flow regimes.

\subsubsection{Quality Control and Preprocessing}
\label{subsubsec:qc_preprocessing}

We apply a multi-tier quality control protocol (detailed in Appendix~\ref{app:qc}) to ensure training data integrity while preserving temporal coverage. Sites with $<90\%$ completeness are excluded, removing 679 stations and yielding 5,321 retained sites from the initial 6,000. Retained sites undergo outlier detection (4$\sigma$ threshold from monthly means), conservative stage--discharge consistency screening~\cite{herschy1999hydrometry}, and temporal consistency validation.

Gap filling uses hierarchical strategies: linear interpolation for gaps $\leq 6$ hours, exponential recession models~\cite{vogel1992baseflow,tallaksen1995hydrograph} for 6-to-24-hour gaps, and exclusion for gaps $>24$ hours. This protocol retains 94.7\% of potential sequences, with 89.3\% containing no interpolation. Windows intersecting flagged spans are dropped, eliminating information leakage.

\subsubsection{Dataset Partitioning}
\label{subsubsec:partitioning}

To assess generalization, we assign entire sites to a single partition, avoiding leakage of site-specific signatures~\cite{kratzert2019towards}. The split allocates 70\% training (n = 3,724), 15\% validation (n = 798), and 15\% test (n = 799), preserving geographic coverage and proportional hydrologic region representation.

Continuous hourly series are segmented into 576-hour windows (24 days), capturing storm hydrographs, weekly patterns, and diurnal fluctuations while remaining tractable for transformers~\cite{vaswani2017attention}. Asymmetric stride (48 hours for training to maximize data utilization through dense overlap, 192 hours for validation and test to reduce autocorrelation) produces 6.03M training sequences, 0.32M validation, and 0.33M test sequences (Table~\ref{tab:dataset_stats}).

\begin{table}[htbp]
\centering
\caption{Foundation corpus statistics}
\label{tab:dataset_stats}
\resizebox{\textwidth}{!}{%
\begin{tabular}{lrrrrr}
\toprule
\textbf{Partition} & \textbf{Sites} & \textbf{Windows (M)} & \textbf{Length (h)} & \textbf{Stride (h)} & \textbf{Timesteps (M)} \\
\midrule
Training & 3,724 & 6.03 & 576 & 48 & 3,470.59 \\
Validation & 798 & 0.32 & 576 & 192 & 185.98 \\
Test & 799 & 0.33 & 576 & 192 & 187.35 \\
\midrule
\textbf{Total} & \textbf{5,321} & \textbf{6.67} & \textbf{576} & --- & \textbf{3,843.92} \\
\bottomrule
\end{tabular}%
}
\end{table}

The held-out test partition provides unseen sites for controlled anomaly benchmarking by injecting anomalies into these sites using the protocol described in Section~\ref{subsubsec:synthetic_test}.

\subsection{Preprocessing Pipeline}
\label{subsec:preprocessing}

\subsubsection{Feature Engineering}
\label{subsubsec:feature_engineering}

At each timestep $t$ for site $s$, the model observes $\mathbf{x}_t \in \mathbb{R}^{12}$ combining static descriptors, dynamic hydrology, and derived temporal context (Table~\ref{tab:features}). This compact design balances information richness, computational tractability, and operational constraints. We intentionally exclude meteorological forcings (precipitation, temperature) to maximize applicability to gauge-only deployment scenarios.

\begin{table}[htbp]
\centering
\caption{12-dimensional feature set composition}
\label{tab:features}
\small
\resizebox{\textwidth}{!}{%
\begin{tabular}{@{}llc@{}}
\toprule
\textbf{Category} & \textbf{Features} & \textbf{Dimension} \\
\midrule
Static basin & Latitude, longitude, drainage area, elevation & 4 \\
Dynamic hydrology & Discharge ($Q$), stage ($H$) & 2 \\
Scale embeddings & $\sigma_{\ln Q}$, $\sigma_{\ln H}$ (training standard deviation of log-transformed series) & 2 \\
Cross-site ranks & Ordinal position: rank($A_d$)/$N$, rank($z_g$)/$N$ & 2 \\
Seasonal context & Monthly anomalies: $(Q - \mu_{Q,m})/\sigma_{Q,m}$, $(H - \mu_{H,m})/\sigma_{H,m}$ & 2 \\
\midrule
\textbf{Total} & & \textbf{12} \\
\bottomrule
\end{tabular}%
}
\end{table}

\textbf{Static descriptors} $\{\phi, \lambda, A_d, z_g\}$ provide coarse hydrograph controls through latitude (eg climatic gradients), longitude (eg continental position), drainage area (eg flashiness), and elevation (eg temperature lapse rates). \textbf{Dynamic variables} $\{Q, H\}$ constitute the core physical state, with stage encoding local geometry and backwater effects~\cite{nittrouer2013backwater} not visible in discharge alone. \textbf{Derived features} include scale embeddings returning absolute variability information lost during standardization, cross-site ranks stabilizing learning across magnitude orders, and monthly anomalies expressing seasonal departures.

\subsubsection{Hierarchical Normalization}
\label{subsubsec:normalization}

Hydrological magnitudes vary by 6 orders across sites (0.1 to 100,000 ft$^3$/s), creating severe optimization challenges. Standard approaches are insufficient: global standardization yields large-river dominance; site-specific standardization alone loses cross-site comparability; min-max scaling is sensitive to outliers; and raw units often prevent stable training.

We introduce a 3-tier hierarchical normalization (full mathematics in Appendix~\ref{app:normalization}) that achieves (1) stable gradients across extreme heterogeneity, (2) no train-test leakage, (3) exact physical unit recovery, and (4) scale-dependent information preservation.

\textbf{Tier 1 Logarithmic stabilization}: Apply $Q^{(1)} = \ln(Q + \epsilon)$ and $H^{(1)} = \ln(H + \epsilon)$ for approximately log-normal variables~\cite{sangal1970lognormal}, which linearizes rating-curve-like relationships and stabilizes variance.

\textbf{Tier 2 Site-specific standardization}: Compute $\mu_s$ and $\sigma_s$ exclusively from each training site series, then transform $\mathbf{x}^{(2)} = (\mathbf{x}^{(1)} - \mu_s)/(\sigma_s + \epsilon)$. For validation and test sites, we apply global training statistics to avoid leakage and to reflect a deployable setting without site-specific calibration.

\textbf{Tier 3 Global clipping}: Apply $\mathbf{x}^{\text{norm}} = \text{clip}(\mathbf{x}^{(2)}, -3, +3)$ to inputs only, retaining 99.7\% of typical variation while preventing gradient explosion. Outputs are denormalized without clipping.

\textbf{Exact inverse}: $\hat{y} = \exp[\hat{y}^{\text{norm}} \cdot (\sigma_s + \epsilon) + \mu_s] - \epsilon$ recovers physical units. Scale embeddings ($\sigma_{\ln Q}$ and $\sigma_{\ln H}$) return absolute variability information, enabling scale-dependent behavior (eg flashiness in small basins).

\textbf{Rationale}: The logarithm converts multiplicative noise to additive; site-specific standardization enables weight sharing without large-river dominance; clipping provides numerical stability; and embeddings distinguish 10 ft$^3$/s from 10,000 ft$^3$/s despite identical normalized values. No single-tier normalization achieves all requirements for continental-scale learning.

\subsection{Training-Time Anomaly Injection}
\label{subsec:anomaly_injection}

Anomaly detection requires labeled corruption patterns, but manual annotation of millions of sequences is prohibitively expensive. We implement on-the-fly synthetic injection that generates diverse patterns during Stage 2 fine-tuning for anomaly detection.

\textbf{Deliberate simplification philosophy}: The training injector implements approximately 11 simplified patterns (spikes, drift, flatlines, dropouts, saturation, clock shifts, quantization, unit jumps, warping, splicing) applied in normalized log-space, deliberately less sophisticated than the test anomalies. This design forces learning of fundamental hydrometric consistency principles (discharge-stage coupling, temporal smoothness, physical plausibility) rather than memorizing specific signatures. If training and test used identical patterns, strong performance could arise from pattern matching rather than genuine understanding. The training-test complexity gap creates a defensible generalization test.

\textbf{Controlled coverage}: A 2-tier system assigns light corruption (60\% probability, 5-to-15\% coverage) or moderate corruption (40\% probability, 15-to-30\% coverage) with iterative refinement ($\pm 3\%$ tolerance for light, $\pm 5\%$ for moderate), maintaining mean 15.2\% $\pm$ 3.1\% across batches. Mixing uses single-type corruption (60\%) to teach type-specific signatures and double-type corruption (40\%) to teach discrimination. Each window uses $n \in [2,4]$ segments with lengths $L \in [T/100, T/4]$ at random positions. Curriculum scheduling ramps injection probability from 0.2 (epochs 1 to 2) to 0.4 thereafter. Complete implementation details appear in Appendix~\ref{app:train_injector}.

\subsubsection{Synthetic Test Set Design}
\label{subsubsec:synthetic_test}

Rigorous generalization assessment requires test data that prevents memorization and emphasizes functional abstraction. We construct a synthetic test set from 799 held-out USGS sites with anomalies spanning 18 types grounded in USGS operational documentation~\cite{rantz1982measurement,kennedy1984discharge,kennedy1983computation,sauer2010stage} and peer-reviewed sensor anomaly studies~\cite{santos2024unsupervised,leigh2019framework,mansanarez2019shift}. Four orthogonal separation axes between training and test (Table~\ref{tab:separation}) ensure strong performance requires learning fundamental hydrometric principles rather than pattern matching.

\paragraph{Rationale for synthetic evaluation.}
The use of synthetic anomaly injection for benchmarking is established practice in time series anomaly detection research (Table~\ref{tab:tsad_benchmarks}). Synthetic injection addresses the scarcity of labeled anomalies in operational archives, enables controlled variation of anomaly characteristics for systematic evaluation, and supports reproducible benchmarking~\cite{paparrizos2022tsbuad,schmidl2022gutentag}. Our approach differs from prior benchmarks in 2 key ways: (1) anomalies are injected into real USGS hydrological time series rather than fully synthetic data, preserving authentic flow dynamics, and (2) injection parameters are grounded in domain-specific operational standards rather than arbitrary choices, ensuring hydrological relevance. This principle of domain-informed synthetic generation has proven effective in other Earth observation contexts, where geo-typical synthetic data tailored to target region characteristics enables generalization without extensive annotation~\cite{song2025synthetic}. Complementarity with operational validation on Canadian data (Section~\ref{sec:canadian}) provides stronger evidence than either approach alone: synthetic evaluation demonstrates learned concepts under controlled conditions, while operational evaluation confirms practical transfer.

\begin{table}[htbp]
\centering
\caption{Comparison of TSAD benchmarks with synthetic or injected anomalies}
\label{tab:tsad_benchmarks}
\footnotesize
\resizebox{\textwidth}{!}{%
\begin{tabular}{@{}llllcl@{}}
\toprule
\textbf{Benchmark} & \textbf{Year} & \textbf{Venue} & \textbf{Domain} & \textbf{\# Types} & \textbf{Injection Method} \\
\midrule
Yahoo S5~\cite{laptev2015yahoo} & 2015 & Webscope & Web KPIs & 2 & Synthetic + real (outliers, changepoints) \\
NAB~\cite{lavin2015nab} & 2015 & ICML AI & IoT or streaming & --- & Real + artificial anomalies \\
SWaT~\cite{mathur2016swat} & 2016 & CySWater & Water treatment & 36 & Cyberattacks injected into real testbed \\
WADI~\cite{ahmed2017wadi} & 2017 & CySWater & Water distribution & 15 & Cyberattacks injected into real testbed \\
Exathlon~\cite{jacob2021exathlon} & 2021 & VLDB & Distributed systems & 6 & Disturbances injected into Spark cluster \\
TODS~\cite{lai2021tods} & 2021 & NeurIPS D\&B & General & 5 & 100\% synthetic (point, shapelet, seasonal, trend) \\
TSB-UAD~\cite{paparrizos2022tsbuad} & 2022 & VLDB & General & 4 & Transformations (noise, smoothing, outliers) \\
GutenTAG~\cite{schmidl2022gutentag} & 2022 & VLDB & General & 10 & Configurable synthetic generator \\
TimeEval~\cite{wenig2022timeeval} & 2022 & VLDB & General & 10 & Toolkit with GutenTAG generator \\
TSAGen~\cite{wang2021tsagen} & 2022 & TNSM & AIOps or KPI & 3+ & Synthetic (season, noise, trend components) \\
MADE~\cite{pimentel2024made} & 2024 & J Hydrology & Hydrology & 2 & Synthetic pulses and trends \\
\midrule
\textbf{Ours} & \textbf{2025} & \textbf{---} & \textbf{Hydrology} & \textbf{18} & \textbf{Injection on real USGS data} \\
\bottomrule
\end{tabular}%
}
\end{table}

\begin{table}[htbp]
\centering
\caption{Four-axis training-test separation strategy}
\label{tab:separation}
\footnotesize
\resizebox{\textwidth}{!}{%
\begin{tabular}{@{}p{2.2cm}p{3.5cm}p{5.5cm}p{4cm}@{}}
\toprule
\textbf{Axis} & \textbf{Training} & \textbf{Test} & \textbf{Purpose} \\
\midrule
Geographic & 3,724 sites & 799 non-overlapping sites & Eliminate site memorization \\
\addlinespace[0.3em]
Mathematical & Linear drift, basic offsets & Exponential, sigmoid, or polynomial variants (3 to 4 per type) & Require functional abstraction \\
\addlinespace[0.3em]
Temporal & 8-to-96-hour segments & Micro (3 to 58 h), meso (7 to 192 h), macro (72 to 520 h) & Probe scale-invariant detection \\
\addlinespace[0.3em]
Parameter & Coverage 10-to-32\%, nominal severity & Coverage trimodal: 3-to-9\% or 32-to-44\% or 44-to-60\%; severity bimodal & Test distribution boundaries \\
\bottomrule
\end{tabular}%
}
\end{table}

\textbf{Geographic separation}: 0 site overlap eliminates memorization of station-specific patterns.
\textbf{Mathematical separation}: Each test anomaly type implements 3 to 4 equation variants (eg drift uses linear, exponential, sigmoid, or polynomial forms) that produce similar visual patterns through different mechanisms.
\textbf{Temporal separation}: Structured duration regimes (micro, meso, macro) span 3 to 520 hours, with macro durations extended to 520 hours to accommodate documented ice persistence in northern river systems~\cite{ashton1986ice}.
\textbf{Parameter separation}: Coverage and severity distributions avoid overlap with training ranges, testing generalization at distribution boundaries.

\paragraph{Anomaly taxonomy and literature grounding.}
The 18 anomaly types (Table~\ref{tab:anomaly_taxonomy}) represent failure modes documented in USGS technical guidance and operational practice. Sensor failures (dropout, flatline, spike) reflect data transmission and equipment malfunctions described in USGS electronic processing standards~\cite{sauer2002standards}. Hydraulic phenomena (backwater, ice backwater, debris effect, sedimentation) follow stage-discharge relationships documented in WSP 2175~\cite{rantz1982measurement} and TWRI 3-A10~\cite{kennedy1984discharge}. Gradual degradation patterns (drift, rating drift, sensor fouling) align with calibration decay and morphological evolution described by Mansanarez et al~\cite{mansanarez2019shift} and the driftR methodology~\cite{shaughnessy2019driftr}. Santos-Fernandez et al~\cite{santos2024unsupervised} report empirical prevalence rates from labeled USGS data: drift (4.26\%), large spikes (0.13\%), small spikes (0.16\%), and flatlines or calibration errors (1.09\%), confirming these failure modes occur in operational archives.

\begin{table}[htbp]
\centering
\caption{Consolidated anomaly taxonomy (18 types)}
\label{tab:anomaly_taxonomy}
\footnotesize
\resizebox{\textwidth}{!}{%
\begin{tabular}{@{}llp{7cm}c@{}}
\toprule
\textbf{Category} & \textbf{Types} & \textbf{Literature Basis} & \textbf{Variants} \\
\midrule
Sensor failures & Dropout, flatline, spike & USGS electronic standards~\cite{sauer2002standards}; Leigh et al~\cite{leigh2019framework} & 3 to 4 each \\
Hydraulic phenomena & Backwater, ice backwater, debris effect, sedimentation & WSP 2175~\cite{rantz1982measurement}; TWRI 3-A10~\cite{kennedy1984discharge} & 3 to 4 each \\
Gradual degradation & Drift, rating drift, sensor fouling & Mansanarez et al~\cite{mansanarez2019shift}; driftR~\cite{shaughnessy2019driftr} & 3 to 4 each \\
Processing errors & Bias step, desync, quantization & USGS time correction standards~\cite{sauer2002standards}; Horner et al~\cite{horner2018impact} & 3 to 4 each \\
Complex artifacts & Splice, noise burst, gate operation & Operational documentation~\cite{rantz1982measurement} & 3 to 4 each \\
\bottomrule
\end{tabular}%
}
\end{table}

\paragraph{Physical-space injection.}
All test anomalies are injected in denormalized physical space to respect discharge-stage coupling. The process involves 3 steps: (1) denormalize to physical units (ft\textsuperscript{3}/s for discharge, ft for stage), (2) apply parametric transformations with hydraulic constraints (eg ice backwater: $H' = H(1 + \alpha_{\text{ice}})$ and $Q' = Q(1 - \beta_{\text{ice}})$ with $\alpha \sim U(0.15, 0.55)$ and $\beta \sim U(0, 0.10)$), and (3) renormalize with clipping. Single-type sequences (30\% of the dataset) enable unambiguous per-type evaluation. Compound anomalies (70\%, with 40\% overlap probability) test discrimination under realistic co-occurrence. Complete injection formulations and reproducibility protocols appear in Appendix~\ref{app:test_injector}.

\begin{figure}[t]
\centering
\includegraphics[width=\textwidth]{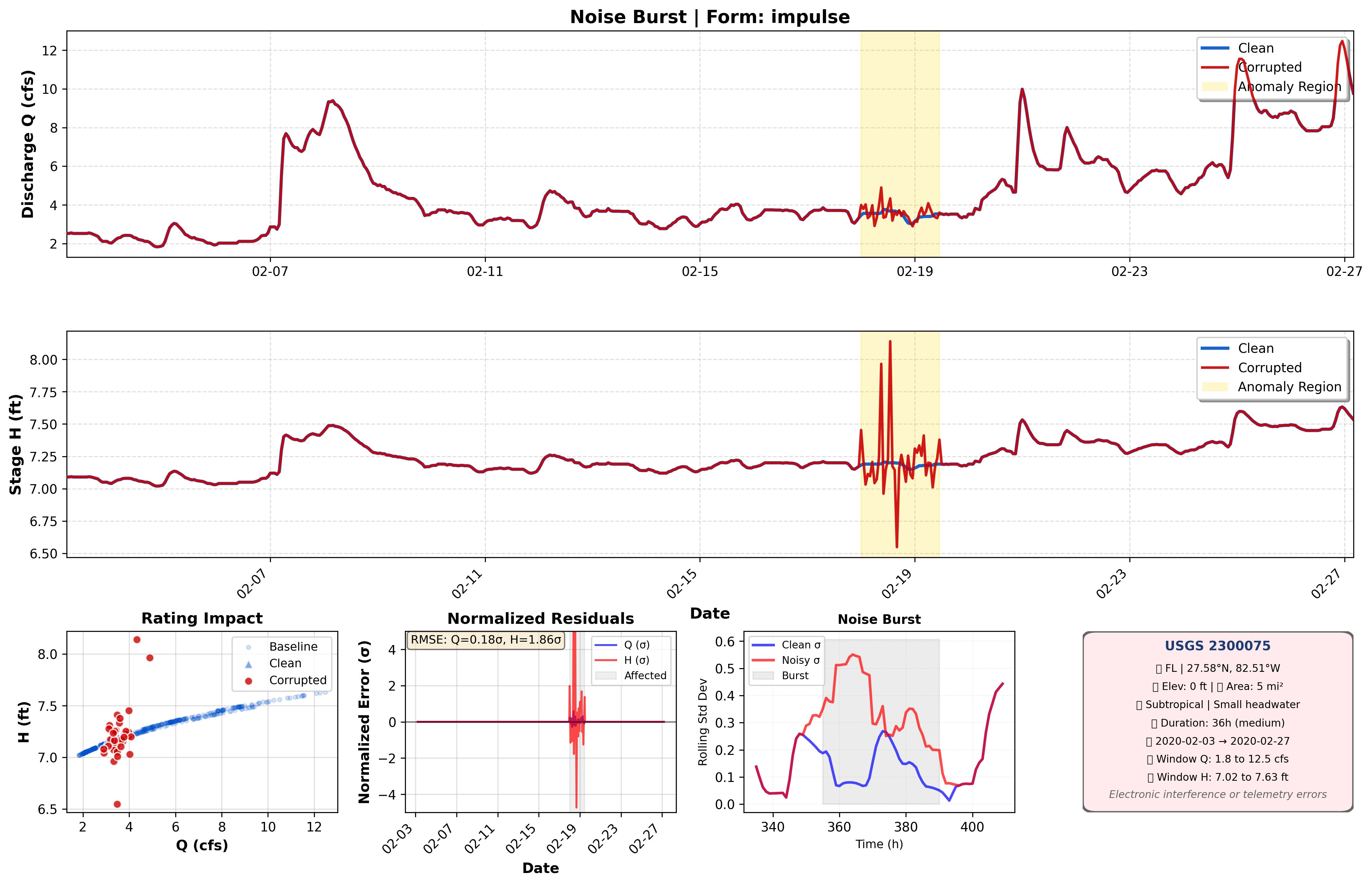}
\caption{Single-segment synthetic injection example illustrating paired clean versus corrupted discharge and stage time series, the affected interval, residual diagnostics, rating-curve impact, and site context.}
\label{fig:synthetic_anomaly_example}
\end{figure}

\paragraph{Anomaly visualization dashboard.}
Figure~\ref{fig:synthetic_anomaly_example} shows a representative single-segment injected anomaly example with paired clean and corrupted discharge and stage signals, residual diagnostics, rating-curve impact, and site context. For additional qualitative examples spanning all 18 anomaly types and equation form variants, see the anomaly visualization dashboard provided in the HydroGEM Hugging Face repository.

\subsubsection{Canadian Zero-Shot Transfer Evaluation}
\label{sec:canadian}

The USGS archive provides quality-controlled products but lacks systematic raw-versus-corrected record pairs that expose operational correction workflows. To test real-world transfer, we evaluate HydroGEM on Environment and Climate Change Canada (ECCC) stations that distribute hourly raw and corrected unit values for both discharge and stage.

\textbf{Zero-shot protocol}: HydroGEM was trained exclusively on USGS sites and never fine-tuned on Canadian data. All experiments are strictly zero-shot, testing whether learned representations generalize across political boundaries, agencies, instrumentation protocols, and rating-curve derivation methods.

\textbf{Data preprocessing}: We obtained ECCC unit value archives, aggregated to hourly resolution, and converted to USGS-aligned imperial units (stage: m to ft via 3.28084; discharge: m$^{3}$/s to cfs via 35.3147). Temporal alignment ensured all 4 series (stage raw, stage corrected, discharge raw, discharge corrected) were available at each hourly timestep.

\textbf{Quality filtering}: To ensure meaningful evaluation, we applied hydrologically-motivated checks to corrected records: sufficient variability (CV $>0.10$), monotonic rating ($\rho > 0.5$), reasonable exponent ($b \in [0.5, 10]$), moderate fit ($R^2 \geq 0.3$), valid data fraction ($\geq 70\%$), and limited flatlines ($<30\%$ of differences $<0.001$ ft). Station-level filtering excluded sites losing $>5\%$ of timesteps. Detailed filtering protocols appear in Appendix~\ref{app:canadian_qc}.

\textbf{Weak label construction}: We compute relative changes induced by operational correction: $\Delta Q_{\text{rel}} = |Q_{\text{corr}} - Q_{\text{raw}}|/(|Q_{\text{raw}}| + \epsilon)$, and similarly for $H$. Timesteps with $\Delta Q_{\text{rel}} > 0.01$ or $\Delta H_{\text{rel}} > 0.01$ are marked anomalous ($y_{\text{corr}} = 1$), providing weak labels that encode where human experts deemed raw data untrustworthy. The 1\% threshold is intentionally conservative, capturing meaningful hydrograph adjustments. These labels indicate where operational corrections occurred rather than providing exhaustive anomaly ground truth. We therefore report tolerant and segment-level metrics in Section~\ref{sec:results}.

These labels are weak because human editors may miss subtle anomalies or apply corrections for reasons orthogonal to sensor data quality (eg re-rating after channel changes). Nevertheless, they provide externally-validated supervision unavailable in synthetic benchmarks. Pattern-based labels (local correlation deviations) complement correction-based labels by emphasizing multi-point segments where corrected hydrograph shape deviates from raw.

\textbf{Site sampling}: From all stations satisfying quality criteria, we randomly sampled 100 distinct stations for evaluation, ensuring representative out-of-sample behavior without cherry-picking. Window filtering required 10-to-40\% correction fraction, identifying substantial but not overwhelming editing effort. For each site, we evaluate HydroGEM in pure zero-shot configuration, comparing predicted anomaly masks against correction-based and pattern-based weak labels.

\section{HydroGEM Model Architecture}
\label{sec:architecture}

\begin{figure*}[htbp]
    \centering
    \includegraphics[width=\textwidth]{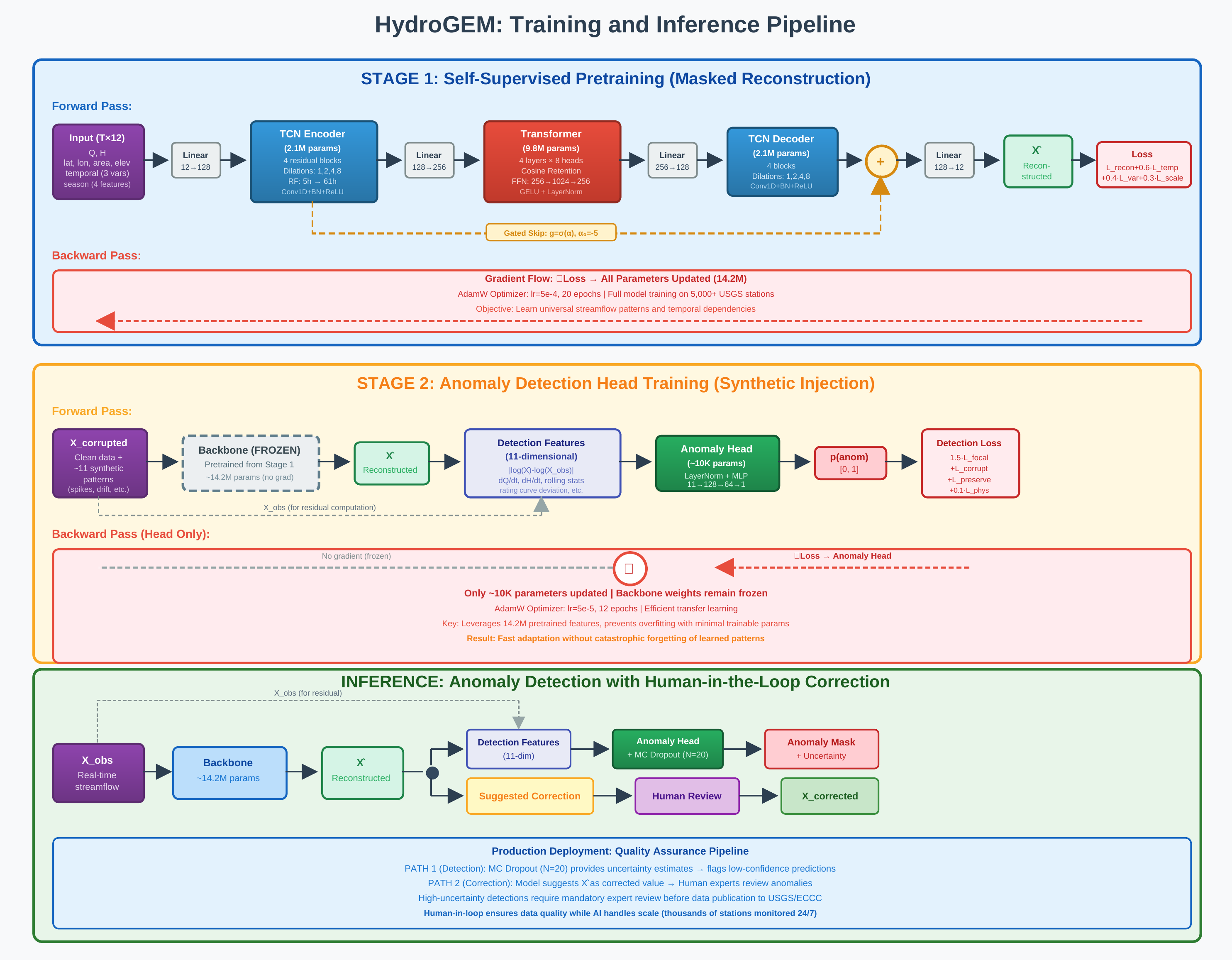}
    \caption{HydroGEM training and inference pipeline. Stage 1 pretrains a hybrid TCN-Transformer backbone (14.2M parameters) using masked reconstruction on clean USGS data. Stage 2 freezes the backbone and trains a lightweight detection head (~10K parameters) on synthetically corrupted sequences. At inference, the model provides anomaly probabilities with uncertainty estimates via MC Dropout, and suggested corrections for human review.}
    \label{fig:architecture}
\end{figure*}

Having established the data preparation pipeline in Section~\ref{sec:data}, we now describe the HydroGEM model that processes the hierarchically normalized 12-dimensional feature vectors to detect anomalies and propose corrections for expert review in streamflow time series (Figure~\ref{fig:architecture}). The model input $\mathbf{X} \in \mathbb{R}^{T \times 12}$ consists of the features defined in Section~\ref{subsubsec:feature_engineering}, transformed through the three-tier normalization scheme (Section~\ref{subsubsec:normalization}), and segmented into 576-hour windows (Section~\ref{subsubsec:partitioning}). Each sequence represents a continental-scale sample spanning diverse hydrologic regimes, requiring the architecture to learn scale-invariant patterns while respecting physical constraints.

\subsection{Two-Stage Training Framework}
\label{subsec:two_stage}

HydroGEM uses a two-stage training approach that decouples representation learning from task-specific optimization, following recent advances in foundation model development~\cite{devlin2019bert,He2022}. This architectural design addresses the challenge of learning robust hydrological representations from limited labeled anomaly data while maintaining generalization across diverse monitoring sites and flow regimes.

The first stage trains a deep autoencoder backbone through self-supervised learning on clean hydrological sequences, allowing the model to discover core relationships between discharge, stage, basin characteristics, and temporal dynamics without requiring explicit anomaly labels. This pretraining phase processes the 6.03 million training sequences described in Section~\ref{subsubsec:partitioning} to develop general-purpose hydrological representations that capture the full spectrum of normal flow behaviors across continental scales. The second stage freezes the pretrained backbone and trains a lightweight detection head using the synthetically generated anomalies described in Section~\ref{subsec:anomaly_injection}, which implements approximately 11 simplified corruption patterns applied in normalized space with controlled coverage (mean 15.2\% $\pm$ 3.1\%).

This decoupled approach provides several advantages over end-to-end anomaly detection architectures. The self-supervised pretraining exploits large volumes of unlabeled data that would be impractical to manually annotate at this scale, learning subtle hydrological patterns that might not be captured from limited labeled examples. The pretrained representations can support transfer to anomaly patterns not explicitly represented in the training injector~\cite{TransferLearning2021}. Furthermore, the modular design permits independent optimization of reconstruction quality and detection performance, facilitating ablation studies and architectural improvements without complete retraining.

\subsection{Stage 1: Self-Supervised Pretraining}
\label{subsec:stage1}

\subsubsection{Backbone Network Architecture}
\label{subsubsec:backbone}

The backbone combines Temporal Convolutional Networks (TCNs) with Transformer-based attention in a hierarchical encoder-decoder structure. This hybrid approach exploits complementary strengths: TCNs for efficient multi-scale local pattern extraction and Transformers for long-range dependencies. Complete architectural equations appear in Appendix~\ref{app:architecture}.

\paragraph{TCN Encoder.}
The encoder maps 12-dimensional inputs to 128-dimensional hidden representations through four stacked TCN blocks with exponentially increasing dilation rates ($r \in \{1, 2, 4, 8\}$). Each block uses residual connections, batch normalization, and dropout ($p = 0.2$). With kernel size $k = 3$, this yields receptive fields spanning roughly 5 to 61 hours across blocks, capturing hydrological processes from sub-daily fluctuations to multi-day storm events. The TCN encoder contributes 2.1 million parameters.

\paragraph{Transformer.}
The TCN output is projected from 128 to 256 dimensions and processed through 4 transformer layers with 8-head attention. We use Cosine Retention Attention with learnable temporal decay~\cite{mongaras2025cottention,sun2023retnet} to improve stability on long sequences. Sliding window attention with window size $W=256$ reduces complexity from $O(T^2)$ to $O(T \cdot W)$. The transformer contributes 9.8 million parameters.

\paragraph{Decoder and Skip Connections.}
The decoder mirrors the encoder, with four TCN blocks reconstructing the original sequence. A gated skip connection adaptively combines encoder and decoder pathways, which can improve gradient flow during early training: the gate parameter $\alpha$ is learned during training, initialized to favor the main pathway. The complete backbone comprises 14.2 million parameters: TCN encoder (2.1M), transformer (9.8M), TCN decoder (2.1M), and projection layers (0.2M).

\subsubsection{Masking Strategy and Objectives}
\label{subsubsec:masking}

The masking strategy in Stage 1 serves a fundamentally different purpose than the anomaly injection in Stage 2 (Section~\ref{subsec:anomaly_injection}). While Stage 2 trains the model to detect corrupted values, Stage 1 establishes a robust representation of normal hydrological behavior through reconstruction of missing data. This pretraining creates a strong inductive bias: the model learns the statistical regularities, physical constraints, and temporal dynamics that characterize clean hydrological data. When subsequently exposed to anomalies during Stage 2, the model can detect them precisely because they violate these learned patterns of normality.

The distinction between missing and corrupted data is critical. Missing data (Stage 1) requires the model to leverage contextual information and physical relationships to infer likely values, teaching it the underlying structure of hydrological systems. Corrupted data (Stage 2) contains incorrect values that must be identified and corrected, a task made possible by the model's pretrained understanding of what constitutes normal behavior. This two-stage approach has proven successful in vision~\cite{He2022} and language understanding~\cite{devlin2019bert}, and we adapt it here for hydrological quality control.

We implement four masking patterns calibrated to the typical data gaps encountered in operational monitoring, informed by masking strategies in self-supervised learning~\cite{Wettig2023}:

\begin{itemize}
\item \textbf{Point masking (40\% probability)}: Randomly masks 15\% of timesteps, forcing reconstruction from immediate temporal context. This pattern addresses telemetry dropouts and transmission noise common in real-time data streams.

\item \textbf{Block masking (30\% probability)}: Masks 1-3 contiguous blocks of 12-72 hours, requiring the model to learn recession curves, diurnal patterns, and multi-day correlations. This pattern reflects sensor outages and extended maintenance windows.

\item \textbf{Periodic masking (20\% probability)}: Masks 4-hour windows at 168-hour intervals, teaching the model to recognize weekly operational cycles. This pattern can reflect scheduled operational patterns and anthropogenic influences where present.

\item \textbf{Feature masking (10\% probability)}: Masks either discharge (70\%) or stage (30\%) for 24-168 hours, forcing the model to learn rating-curve relationships~\cite{kennedy1984discharge} and cross-variable dependencies essential for detecting coupled sensor failures.
\end{itemize}

Each training sequence undergoes masking with probability 0.80. The model learns to reconstruct discharge and stage (the primary targets), while other features provide contextual information. The resulting reconstruction task requires the model to internalize three critical aspects of hydrological data: (1) temporal dynamics including recession rates and hydrograph shapes, (2) physical relationships between discharge and stage governed by hydraulic geometry, and (3) scale-dependent patterns that vary with basin size and flow regime. These learned representations provide the foundation for detecting anomalies that violate expected patterns during operational deployment.

\subsubsection{Pretraining Loss Functions}
\label{subsubsec:pretrain_loss}

The pretraining objective combines five loss components (complete formulations in Appendix~\ref{app:losses}):

\begin{itemize}
\item \textbf{Weighted reconstruction loss}: Prioritizes discharge (weight 3.0) and stage (2.5) over other features
\item \textbf{Temporal consistency loss}: Preserves recession slopes and rising limb characteristics
\item \textbf{Variance preservation loss}: Prevents over-smoothing to collapsed mean predictions
\item \textbf{Scale consistency loss}: Maintains denormalization capability for physical unit recovery
\item \textbf{Attention regularization}: Encourages diverse attention patterns across heads
\end{itemize}

The backbone trains for 20 epochs using AdamW optimization~\cite{Loshchilov2017} with OneCycleLR scheduling (peak learning rate $5 \times 10^{-4}$), gradient clipping (norm 1.0), and early stopping (patience 7 epochs). Training requires approximately 48 GPU-hours on NVIDIA A100 hardware.

\subsection{Stage 2: Anomaly Detection Head Training}
\label{subsec:stage2}

The second stage freezes the pretrained backbone and trains a lightweight detection head using the synthetic anomaly injection framework described in Section~\ref{subsec:anomaly_injection}. As detailed in that section, the training injector implements approximately 11 simplified anomaly patterns (spikes, drift, flatlines, dropouts, saturation, clock shifts, quantization, unit jumps, warping, splicing, and subtle drift) applied in normalized log-space with controlled coverage targeting 15.2\% $\pm$ 3.1\% temporal coverage. This deliberate simplification forces the model to learn fundamental hydrometric consistency principles rather than memorizing specific anomaly signatures: simple corruption mechanisms are used during training while testing employs complex physical-space anomalies (Section~\ref{subsubsec:synthetic_test}).

\subsubsection{Detection Head Architecture}
\label{subsubsec:detection_head}

The detection head operates entirely on observable quantities without requiring ground-truth clean data during inference~\cite{Angiulli2023,Gong2019}. It computes an 11-dimensional feature vector from the relationship between potentially corrupted observations and backbone reconstructions. These features are chosen to reflect operationally observable inconsistencies between the observations and the model's reconstruction, without requiring access to a ground-truth clean series:

\begin{itemize}
\item \textbf{Reconstruction residuals}: Absolute differences between observed and reconstructed discharge and stage
\item \textbf{Temporal gradients}: Forward differences revealing drift and trend anomalies
\item \textbf{Rolling variability}: 7-hour window statistics distinguishing sensor malfunction from natural variability
\item \textbf{Rating-curve deviation}: Local power-law fit quality indicating hydraulic inconsistency
\item \textbf{Cross-correlation features}: Coupled anomalies affecting both variables simultaneously
\end{itemize}

Features undergo robust standardization using median absolute deviation and are processed through a two-layer MLP (128 and 64 units) with GELU activation and dropout ($p = 0.2$), yielding approximately 10K trainable parameters. Dropout is enabled during training; at inference we use Monte Carlo dropout with N=20 stochastic forward passes to estimate predictive uncertainty.

\subsubsection{Detection Head Training Objectives}
\label{subsubsec:finetune_loss}

The detection head training loss balances four objectives (complete formulations in Appendix~\ref{app:losses}):

\begin{itemize}
\item \textbf{Focal loss}: Addresses class imbalance with $\alpha = 0.25$, $\gamma = 2.0$
\item \textbf{Corruption reconstruction}: Ensures accurate corrections on anomalous segments
\item \textbf{Clean preservation}: Penalizes modifications to uncorrupted data (essential for operational trust)
\item \textbf{Physics constraints}: Penalizes discharge-stage inconsistencies (e.g., deviation from local rating-curve fits) and discourages physically implausible corrections
\end{itemize}

Training proceeds for 12 epochs with AdamW ($lr = 5 \times 10^{-5}$), requiring approximately 8 GPU-hours. Because the backbone remains frozen, only the ~10K detection head parameters are updated, enabling efficient adaptation while preserving the pretrained backbone representations.

\subsection{Inference Pipeline}
\label{subsec:inference}

At deployment, the trained model processes incoming observations through the backbone to generate reconstructions, then computes detection features from observation-reconstruction residuals (Figure~\ref{fig:architecture}, bottom). The inference pipeline produces three outputs per timestep:

\begin{itemize}
\item \textbf{Anomaly probability}: A score in [0, 1] indicating likelihood of data quality issues
\item \textbf{Uncertainty estimate}: Standard deviation across MC Dropout passes (N=20), flagging low-confidence predictions for mandatory review
\item \textbf{Suggested corrections}: Reconstructed discharge ($\hat{Q}$) and stage ($\hat{H}$) values that require hydrologist approval before integration into official records
\end{itemize}

A three-tier decision system operationalizes these outputs: (1) high-confidence clean predictions (low anomaly probability, low uncertainty) pass with minimal review; (2) high-confidence anomaly detections trigger automatic flagging with suggested corrections; (3) uncertain predictions (high uncertainty regardless of probability) require mandatory expert inspection. Thresholds for the three tiers are selected on the validation set and reported in Section~\ref{sec:results}. This human-in-the-loop design ensures data quality while enabling continuous monitoring across thousands of stations.


\section{Experimental Setup}
\label{sec:experiments}

\subsection{Computational Infrastructure}
\label{subsubsec:compute}

All experiments were run on the Lonestar6 supercomputer at the Texas Advanced Computing Center, accessed through the NSF-led National AI Research Resource (NAIRR) Pilot program~\cite{tacc2024nairr_lonestar6,nsf2024nairrpilot}. Stage 1 pretraining ran for 20 epochs and required approximately 48 GPU-hours on a single NVIDIA A100 GPU; Stage 2 detection head training ran for 12 epochs on a single NVIDIA A100 GPU. We used mixed-precision training with PyTorch and deterministic seeds fixed per experiment. Unless otherwise stated, we select checkpoints using the lowest validation loss (from clean validation sequences only) on the USGS validation partition (n=798 sites; Section~\ref{subsubsec:partitioning}), and we never use the USGS test partition or any Canadian station for model selection.

\subsection{Evaluation Metrics}
\label{subsec:metrics}

We evaluate HydroGEM along two complementary dimensions: anomaly detection performance and reconstruction quality.

For anomaly detection we report precision, recall, and F1 score. F1 is our primary reporting metric because it matches the operational setting, where false alarms and missed anomalies both carry cost under a fixed decision rule. Unless otherwise stated, binary predictions are obtained by thresholding anomaly probabilities at 0.5, which we use as a fixed default threshold for all experiments to avoid post hoc tuning.

For reconstruction quality we focus on settings where a clean reference is available. On the synthetic USGS test set we compute the relative reduction in absolute error between raw anomalous values and model reconstructions,
\begin{equation}
\text{Error Reduction} = 
\frac{\lvert X^{(\text{raw})} - X^{(\text{clean})} \rvert - \lvert \hat{X} - X^{(\text{clean})} \rvert}
{\lvert X^{(\text{raw})} - X^{(\text{clean})} \rvert + \epsilon} \times 100\%,
\end{equation}
where $X \in \{Q, H\}$ and $\epsilon$ is a small constant for numerical stability. We compute metrics separately for discharge and stage and report segment-level averages together with root mean squared error (RMSE) on anomalous segments. We additionally track RMSE on non-anomalous timesteps to confirm that the model does not degrade clean observations. Error reduction and RMSE are computed after denormalization in physical units, restricted to injected anomalous intervals for anomalous-segment metrics.

For the Canadian case study, reconstructions are evaluated primarily as a diagnostic. Weak labels are derived from operational corrections as described in Section~\ref{sec:canadian}, and correction strategies differ between USGS and Environment and Climate Change Canada, particularly for ice-affected periods~\cite{usgs_wdr_docs,eccc_data_manual,eccc_ice_warning}. As a result, we treat reconstruction error as a secondary diagnostic rather than a primary performance metric and place emphasis on detection metrics.

Because the Canadian weak labels are derived from operational corrections recorded with daily granularity rather than precise anomaly boundaries, we report two complementary detection metrics. First, we report pointwise F1, where predictions and labels are compared at each hourly timestep. Second, we report tolerant F1 with a $\pm$24-hour buffer, which credits a prediction if it falls within $\pm 24$ hourly timesteps of a labeled correction timestep. In addition, we report segment-level recall to evaluate event detection under boundary uncertainty: contiguous labeled correction intervals are treated as anomaly events and an event is counted as detected if any predicted anomaly overlaps the interval or falls within the tolerance window. This multi-metric approach follows recommendations for evaluating time-series anomaly detection under label uncertainty~\cite{kim2022rigorous,tatbul2018precision}.

\subsection{Baseline Methods}
\label{subsec:baselines}

To assess zero-shot anomaly detection performance we compare HydroGEM against 11 baseline methods that do not require labeled training data or site-specific calibration. Specifically, we include 3 statistical baselines (Z-Score, IQR, Moving Average Residual), 3 generic unsupervised baselines (Isolation Forest, LOF, STL residual), and 5 hydrology-motivated baselines (rating-curve residual, rate-of-change limits, persistence detection, $Q$--$H$ consistency checks, and a seasonal envelope). All baselines operate in a zero-shot mode at new stations, matching HydroGEM's deployment setting. Appendix~\ref{app:baselines} contains full baseline parameters and implementation details. For fair comparison, all baselines operate on the same hourly discharge and stage series and use fixed hyperparameters shared across all stations. We consider three categories.

\paragraph{Statistical baselines.}

These methods represent simple distributional checks that are widely used in practice. They include a Z-Score detector that flags observations with absolute z-score greater than 3 for either discharge or stage, an Interquartile Range (IQR) rule that flags values outside the interval $[Q_1 - 1.5 \times \text{IQR}, Q_3 + 1.5 \times \text{IQR}]$, and a Moving Average Residual detector that uses a centered 7-day window and flags points whose residuals exceed 3 standard deviations of the rolling window. Together these capture pointwise outliers and deviations from local temporal context.

\paragraph{Unsupervised machine learning.}

We include generic unsupervised anomaly detectors that require no labels and can be applied at scale. Isolation Forest and Local Outlier Factor (LOF) operate on standardized discharge and stage and use fixed contamination and neighborhood settings specified in Appendix~\ref{app:baselines}, representing tree-based and density-based approaches to unsupervised outlier detection. A Seasonal Trend decomposition using LOESS (STL) baseline models a 7-day seasonal component and flags residuals that exceed 3 standard deviations. These methods provide a representative sample of modern off-the-shelf anomaly detectors that do not use hydrological structure.

\paragraph{Hydrological domain methods.}

The third group encodes relationships that hydrologists routinely use during manual quality control. A Rating Curve Residual baseline fits a power-law relationship $Q = a (H - H_0)^b$ by log-linear regression and flags points with large log-space residuals. This serves as a key reference because rating residuals are central to operational practice for detecting rating shifts, drift, and ice effects. Additional checks include rate-of-change limits on discharge and stage, persistence detection for stuck sensors, $Q$--$H$ consistency checks that flag periods where discharge and stage trends diverge or correlations break down, and a seasonal envelope that flags values outside month-specific percentile bands. Together these methods approximate the mix of physical rules and heuristics that agencies use in production systems.

\paragraph{Methods not included.}

We intentionally exclude methods that rely on extensive labeled anomalies or site-specific tuning. Supervised deep learning classifiers and autoencoders require labels for each station and would not represent a true zero-shot setting. Per-site reconstruction models that are fit directly on test windows are also excluded, since they would not be deployed in this way operationally. Proprietary automated flagging systems from agencies such as USGS and Environment and Climate Change Canada are not reproducible without internal configuration parameters. Pretraining multiple alternative deep architectures on the full USGS corpus would be methodologically interesting but is beyond the present computational scope and is left for future work.

\subsection{Evaluation Protocols}
\label{subsec:protocols}

\paragraph{Synthetic USGS evaluation.}

For synthetic evaluation we use the held-out set of 799 USGS sites described in Section~\ref{subsubsec:synthetic_test} that are completely unseen during training. Clean sequences from these sites are corrupted with injected anomalies that cover a broad range of types, durations, spatial coverage, and severities. We report aggregate detection metrics over all anomalies and also stratify performance by anomaly type and basic regime attributes such as duration and coverage. Detailed breakdowns are provided in the appendix, while the main text focuses on overall detection performance and the relative ranking of HydroGEM against baselines.

Reconstruction quality on this synthetic set is summarized by the error reduction and RMSE metrics described above, computed only on timesteps where injected anomalies are present, with separate reporting of RMSE on clean timesteps.

\paragraph{Canadian zero-shot evaluation.}

For cross-national evaluation we apply HydroGEM without retraining to 100 Canadian stations from Environment and Climate Change Canada. Weak labels are derived from human corrections in the operational record as described in Section~\ref{sec:canadian}. These labels reflect editorial decisions and local correction practice, so we report tolerant F1 (with $\pm$24-hour buffer) as the primary metric, alongside pointwise F1 and segment-level recall, over a range of weak-label thresholds defined by the relative magnitude of human edits, with full curves in the appendix. This provides a sensitivity view of how performance changes as the notion of meaningful correction becomes stricter. Given the documented differences between USGS and Environment and Climate Change Canada correction strategies in ice-affected periods~\cite{usgs_wdr_docs,eccc_data_manual,eccc_ice_warning}, we interpret reconstruction errors here as supportive evidence rather than a primary outcome and focus our conclusions on detection performance.


\section{Results}
\label{sec:results}

We evaluate HydroGEM through two complementary assessments: (1) detection and reconstruction on a synthetic test set with exact ground-truth labels, and (2) zero-shot transfer to Canadian stations where labels derive from operational hydrologist corrections. The synthetic evaluation establishes baseline capabilities under controlled conditions, while the Canadian evaluation tests generalization to real-world data quality challenges across national boundaries.

\subsection{Synthetic Anomaly Detection}
\label{subsec:synthetic_results}

\subsubsection{Overall Detection Performance}
\label{subsubsec:synthetic_overall}

\begin{figure}[htbp]
    \centering
    \includegraphics[width=\columnwidth]{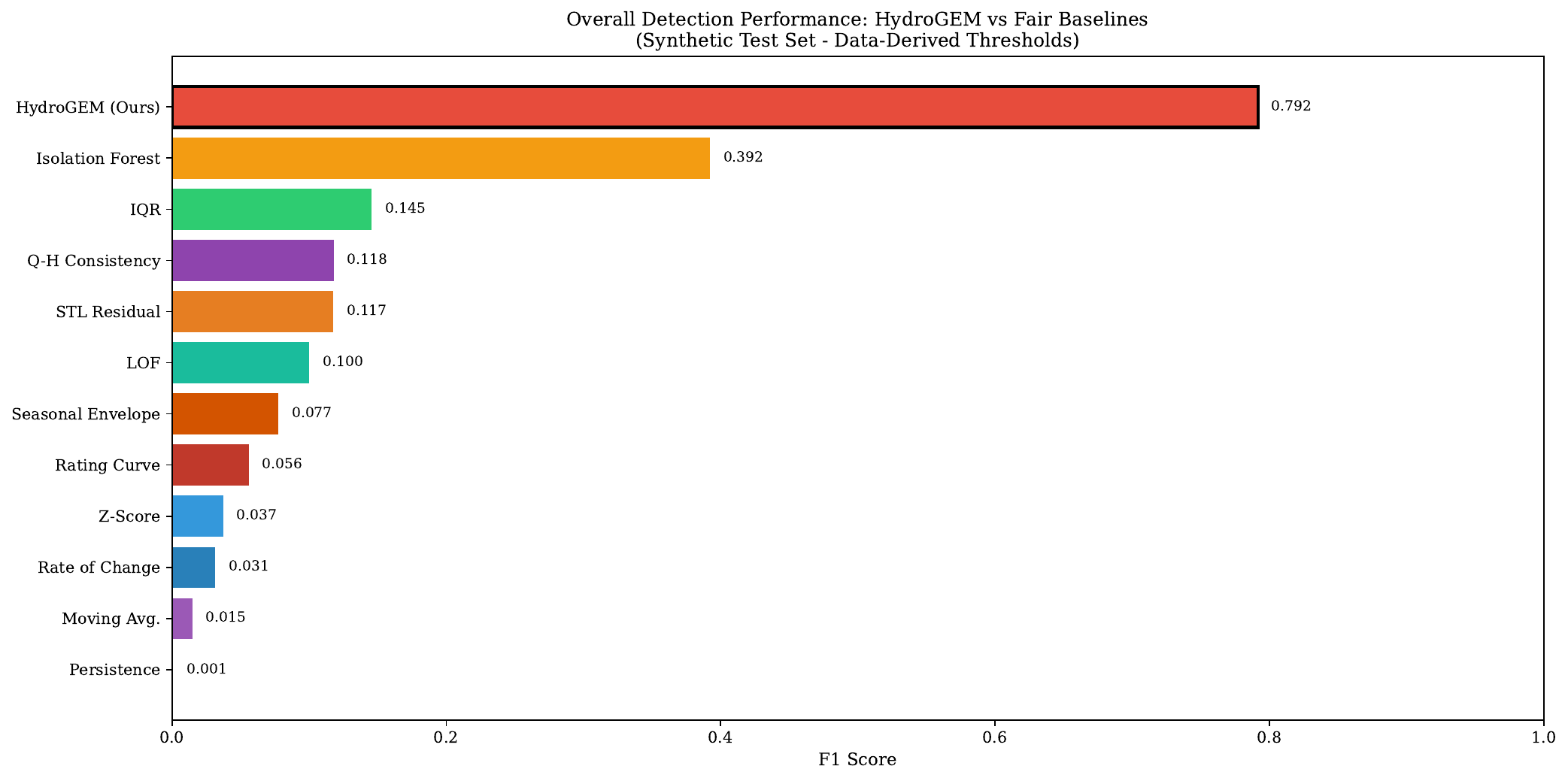}
    \caption{Overall detection performance on the synthetic test set. HydroGEM (F1 = 0.792) outperforms all baselines, with Isolation Forest (F1 = 0.392) as the nearest competitor.}
    \label{fig:synthetic_overall}
\end{figure}

Table~\ref{tab:synthetic_baseline} and Figure~\ref{fig:synthetic_overall} present detection performance for HydroGEM and eleven baseline methods on the synthetic test set. HydroGEM achieves F1 = 0.792 (precision: 0.755, recall: 0.832), substantially outperforming all baselines. The strongest baseline, Isolation Forest, attains F1 = 0.392, so HydroGEM provides an absolute F1 gain of 0.400 (approximately 2.0$\times$ higher F1). Across sites, HydroGEM achieves higher site-level F1 than each baseline (paired Wilcoxon signed-rank test over 799 test sites, $p < 0.001$).

\begin{table}[htbp]
    \centering
    \caption{Detection performance on the synthetic test set. All baseline hyperparameters and thresholds are fixed a priori (Appendix~\ref{app:baselines}) and are not tuned against synthetic test labels.}
    \label{tab:synthetic_baseline}
    \small
    \begin{tabular}{@{}llccc@{}}
        \toprule
        Method & Category & F1 & Precision & Recall \\
        \midrule
        \textbf{HydroGEM (Ours)} & Foundation Model & \textbf{0.792} & 0.755 & 0.832 \\
        \midrule
        Isolation Forest & Unsupervised ML & 0.392 & 0.600 & 0.291 \\
        IQR & Statistical & 0.146 & 0.621 & 0.082 \\
        Q--H Consistency & Hydrological & 0.118 & 0.608 & 0.065 \\
        STL Residual & Unsupervised ML & 0.117 & 0.747 & 0.064 \\
        LOF & Unsupervised ML & 0.100 & 0.563 & 0.055 \\
        Seasonal Envelope & Hydrological & 0.078 & 0.649 & 0.041 \\
        Rating Curve & Hydrological & 0.056 & 0.703 & 0.029 \\
        Z-Score & Statistical & 0.037 & 0.665 & 0.019 \\
        Rate of Change & Hydrological & 0.032 & 0.767 & 0.016 \\
        Moving Average & Statistical & 0.015 & 0.645 & 0.008 \\
        Persistence & Hydrological & 0.001 & 0.004 & 0.001 \\
        \bottomrule
    \end{tabular}
\end{table}

The baseline hierarchy reveals distinct performance tiers. Isolation Forest (F1 = 0.392) achieves the strongest baseline performance by leveraging multivariate structure in the discharge-stage feature space. Traditional statistical methods (Z-Score, IQR, Moving Average) achieve F1 $<$ 0.15, as they operate on univariate signals and cannot capture stage-discharge coupling violations. Domain-specific approaches like Q--H Consistency (F1 = 0.118) and Rating Curve residuals (F1 = 0.056) underperform despite their hydrological grounding---these methods detect only anomalies that violate instantaneous rating relationships, missing temporally extended distortions that preserve local Q--H ratios.

\subsubsection{Per-Anomaly-Type Analysis}
\label{subsubsec:synthetic_pertype}

\begin{figure}[htbp]
    \centering
    \includegraphics[width=\columnwidth]{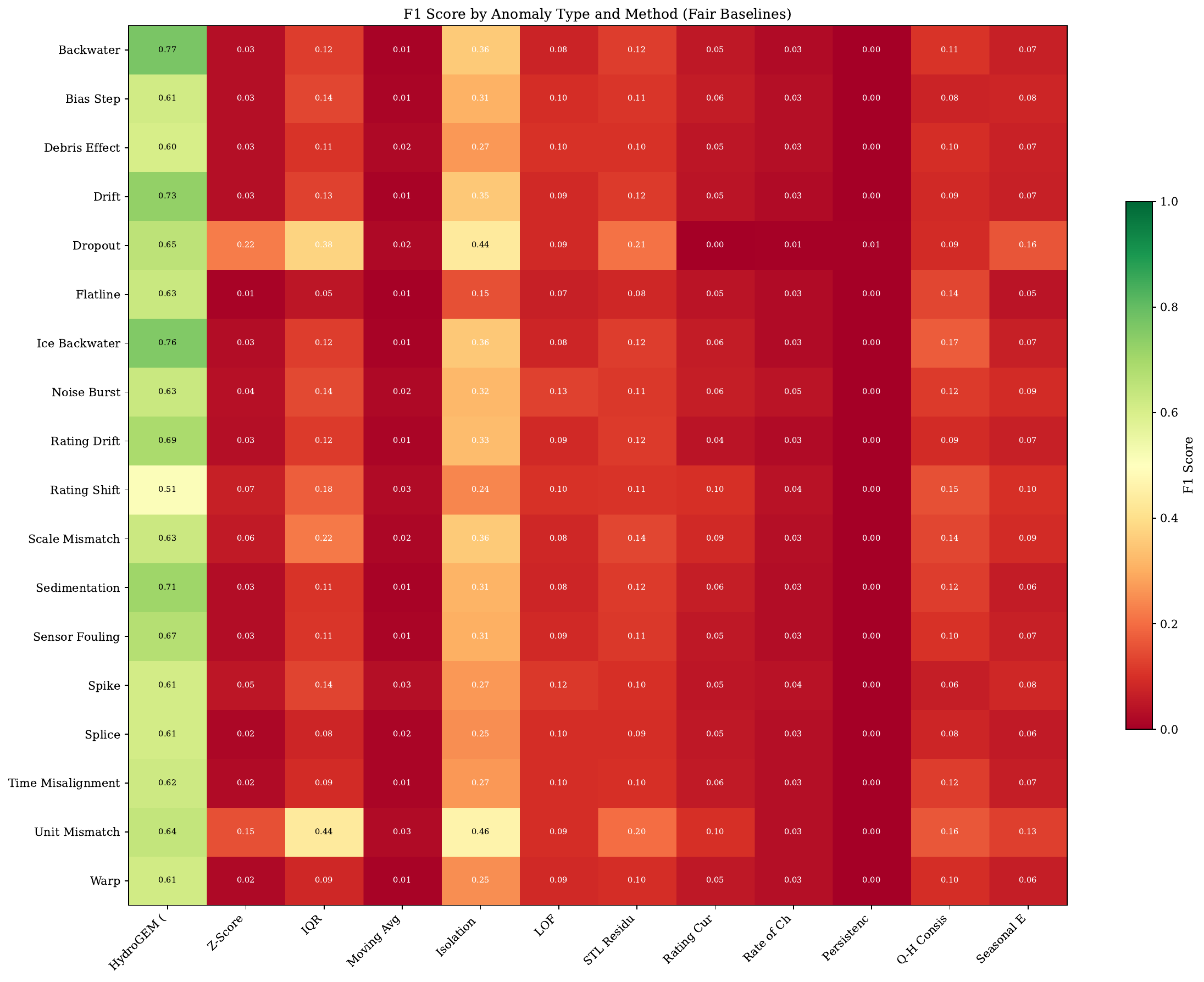}
    \caption{Detection F1 scores by anomaly type and method. HydroGEM achieves the highest performance across all 18 anomaly types. Isolation Forest provides the strongest baseline overall, particularly for dropout and unit mismatch detection.}
    \label{fig:synthetic_heatmap}
\end{figure}

Figure~\ref{fig:synthetic_heatmap} presents F1 scores stratified by anomaly type and detection method. HydroGEM achieves the highest F1 for all 18 anomaly types, demonstrating consistent generalization across the full spectrum of hydrometric failures.

Performance varies systematically with anomaly characteristics. HydroGEM achieves strongest detection for backwater effects (F1 = 0.77), ice backwater (F1 = 0.76), and drift anomalies (F1 = 0.73)---conditions that produce sustained Q--H decoupling patterns. Moderate performance occurs for sensor-level artifacts including bias steps (F1 = 0.61), flatlines (F1 = 0.63), and spikes (F1 = 0.61), which manifest as localized discontinuities rather than extended temporal patterns. The most challenging anomaly type is rating shift (F1 = 0.51), where abrupt but physically plausible changes in the stage-discharge relationship can resemble legitimate rating curve updates.

The heatmap also reveals baseline specializations. Isolation Forest achieves its best performance on dropout (F1 = 0.44) and unit mismatch (F1 = 0.46), both of which produce outliers in multivariate feature space. However, no baseline achieves F1 $>$ 0.20 for subtle anomalies like sensor fouling, splice artifacts, or time misalignment---patterns that require learning complex temporal dependencies rather than applying fixed decision rules.

\subsubsection{Reconstruction Quality}
\label{subsubsec:synthetic_reconstruction}

Beyond detection, HydroGEM provides suggested corrections through its reconstruction output. On the synthetic test set, the model achieves 68.7\% mean error reduction computed using the metric defined in Section~\ref{subsec:metrics}.

Reconstruction performance correlates with anomaly severity and duration. For high-magnitude anomalies ($>$25\% deviation from clean values), error reduction reaches 74.2\%. Medium-duration events (6--48 hours) achieve 71.3\% reduction, compared to 64.1\% for short events ($<$6 hours) where limited temporal context constrains inference. The model preserves clean data with high fidelity: on uncorrupted segments, reconstruction MAE remains below 2\% of the signal range, confirming that corrections are applied selectively.

\subsubsection{Detection Examples}
\label{subsubsec:synthetic_examples}

\begin{figure}[htbp]
    \centering
    \includegraphics[width=\columnwidth]{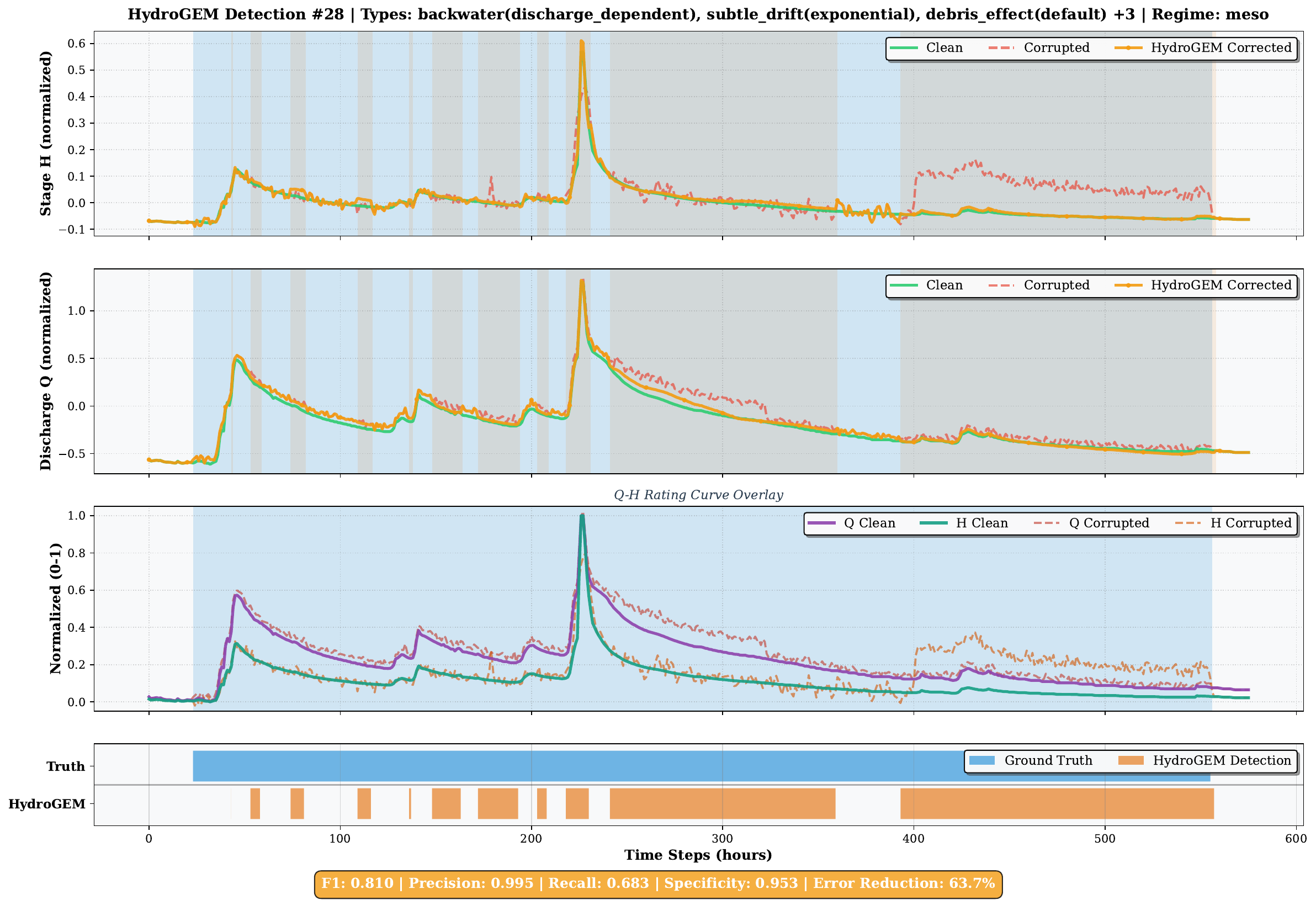}
    \caption{Successful detection example with overlapping backwater, exponential drift, and debris effects. HydroGEM achieves F1 = 0.810 with 63.7\% error reduction. Top panels show stage and discharge time series (green: clean, red dashed: corrupted, orange: HydroGEM corrected). Bottom panel compares ground-truth labels with model detections.}
    \label{fig:synthetic_detection_success}
\end{figure}

Figure~\ref{fig:synthetic_detection_success} illustrates successful detection on a challenging multi-anomaly window containing overlapping backwater effects, exponential drift, and debris-induced artifacts spanning approximately 400 hours. HydroGEM achieves F1 = 0.810 (precision: 0.995, recall: 0.683) with 63.7\% error reduction. The high precision indicates minimal false alarms on clean segments, while the reconstruction closely tracks the ground-truth signal through complex, overlapping corruption patterns.

\begin{figure}[htbp]
    \centering
    \includegraphics[width=\columnwidth]{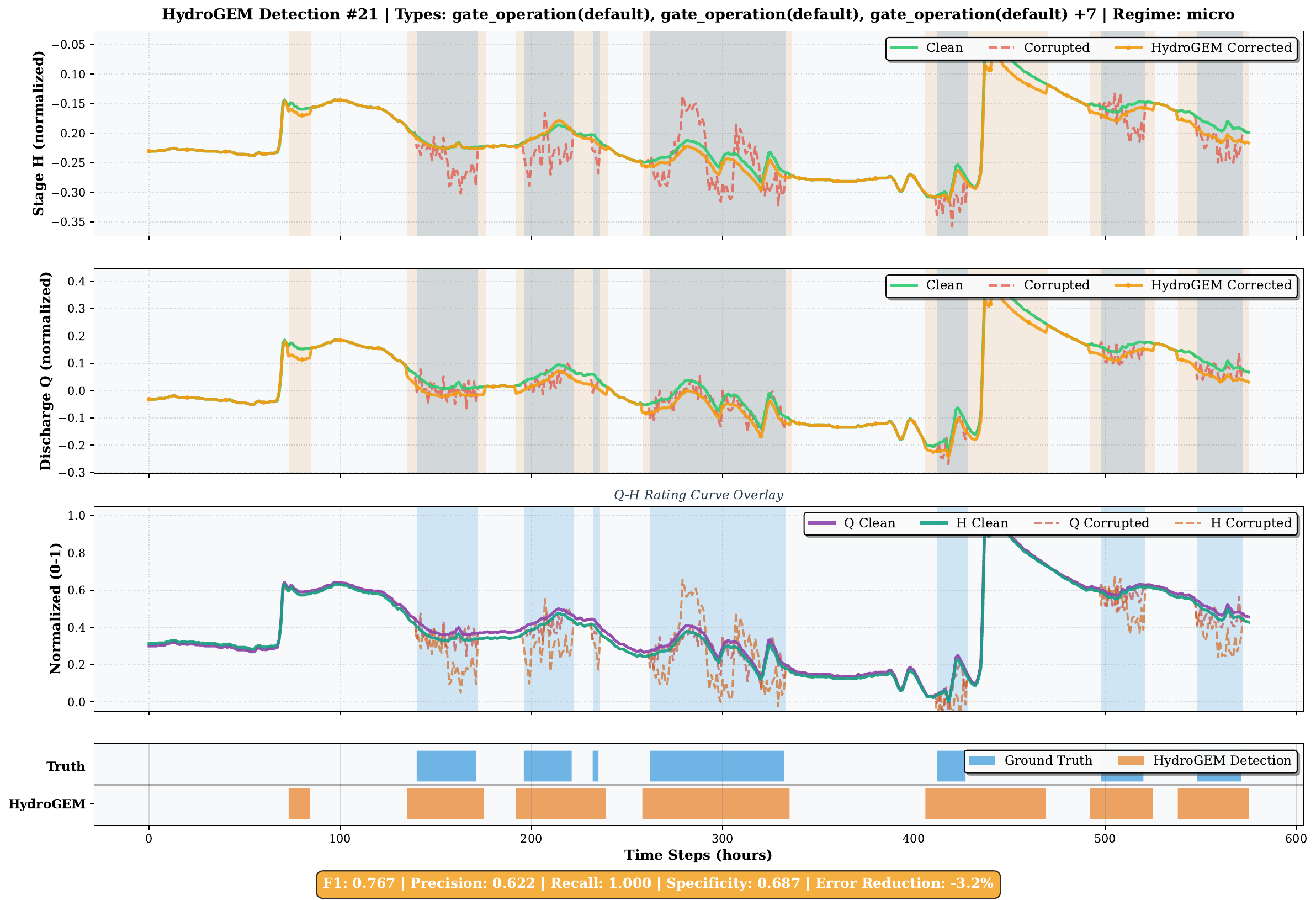}
    \caption{Challenging case with gate operation events. HydroGEM achieves F1 = 0.767 with high recall but reduced precision, as legitimate gate-induced transients produce Q--H patterns similar to sensor artifacts. This motivates treating model outputs as suggestions requiring expert review.}
    \label{fig:synthetic_detection_challenge}
\end{figure}

Figure~\ref{fig:synthetic_detection_challenge} presents a challenging case containing multiple gate operation events producing rapid, physically valid discharge fluctuations. HydroGEM achieves recall of 1.0 but lower precision (0.622), resulting in F1 = 0.767 with slightly negative error reduction ($-$3.2\%). The model flags legitimate gate-induced transients as suspicious---these rapid Q--H excursions resemble sensor artifacts in the learned representation. This case illustrates the inherent difficulty of distinguishing anthropogenic flow modifications from sensor malfunctions without explicit metadata about control structures.

This failure mode motivates an important operational consideration: HydroGEM outputs should be treated as quality control suggestions rather than autonomous corrections, particularly for stations with known flow regulation. The model's high recall ensures that genuine anomalies are rarely missed, while expert review can filter false positives arising from unusual but valid hydrological conditions.

\subsection{Zero-Shot Transfer to Canadian Stations}
\label{subsec:canadian_results}

We evaluated HydroGEM on 100 stations from Environment and Climate Change Canada's (ECCC) hydrometric network, data the model never encountered during training. This assessment tests whether representations learned from USGS stations transfer across national boundaries, instrumentation practices, and climatic regimes.

\subsubsection{Baseline Comparison}
\label{subsubsec:canadian_baseline}

Table~\ref{tab:canadian_baseline} presents pointwise detection performance on Canadian stations. We applied the same eleven baseline methods using identical threshold protocols described in Section~\ref{subsubsec:synthetic_overall}. HydroGEM achieves pointwise F1 = 0.582, substantially outperforming all baselines (paired Wilcoxon signed-rank test over 100 sites, $p < 0.001$).

\begin{table}[htbp]
\centering
\caption{Pointwise detection performance on Canadian zero-shot evaluation (1\% correction threshold).}
\label{tab:canadian_baseline}
\small
\begin{tabular}{@{}llccc@{}}
\toprule
Method & Category & F1 & Precision & Recall \\
\midrule
\textbf{HydroGEM} & Foundation Model & \textbf{0.582} & 0.650 & 0.567 \\
\addlinespace
Persistence & Hydrological & 0.418 & 0.428 & 0.410 \\
Isolation Forest & Unsupervised ML & 0.263 & 0.354 & 0.210 \\
IQR & Statistical & 0.104 & 0.139 & 0.086 \\
LOF & Unsupervised ML & 0.101 & 0.096 & 0.107 \\
STL Residual & Unsupervised ML & 0.058 & 0.111 & 0.039 \\
Seasonal Envelope & Hydrological & 0.049 & 0.082 & 0.035 \\
Rate of Change & Hydrological & 0.038 & 0.067 & 0.027 \\
Rating Curve & Hydrological & 0.028 & 0.036 & 0.034 \\
Z-Score & Statistical & 0.028 & 0.031 & 0.032 \\
Q--H Consistency & Hydrological & 0.024 & 0.022 & 0.027 \\
Moving Average & Statistical & 0.009 & 0.012 & 0.007 \\
\bottomrule
\end{tabular}
\end{table}

Among baselines, persistence detection performs best (F1 = 0.418), reflecting that stuck sensors constitute a meaningful fraction of operational corrections in the Canadian archive. However, persistence detection is narrowly specialized: it identifies only zero-variance intervals and cannot detect rating curve violations, ice effects, sensor drift, or backwater conditions. Isolation Forest ranks second (F1 = 0.263), consistent with its general-purpose outlier detection capability. Statistical methods and most hydrological heuristics achieve F1 $<$ 0.11, indicating that simple threshold-based rules inadequately capture the complexity of real-world data quality issues.

The baseline ranking differs notably between synthetic and Canadian evaluations. Isolation Forest performs strongly on synthetic data (F1 = 0.392) but less so on Canadian data (F1 = 0.263), while Persistence shows the opposite pattern (synthetic: F1 = 0.001; Canadian: F1 = 0.418). This divergence reflects differences in anomaly composition: synthetic anomalies include diverse corruption patterns where multivariate outlier detection excels, whereas Canadian operational corrections predominantly address sensor stalls and ice effects that Persistence can partially identify. HydroGEM's consistent strong performance across both evaluation settings demonstrates robustness to these distributional differences.

\subsubsection{Detection Example}
\label{subsubsec:canadian_example}

\begin{figure}[htbp]
    \centering
    \includegraphics[width=\textwidth]{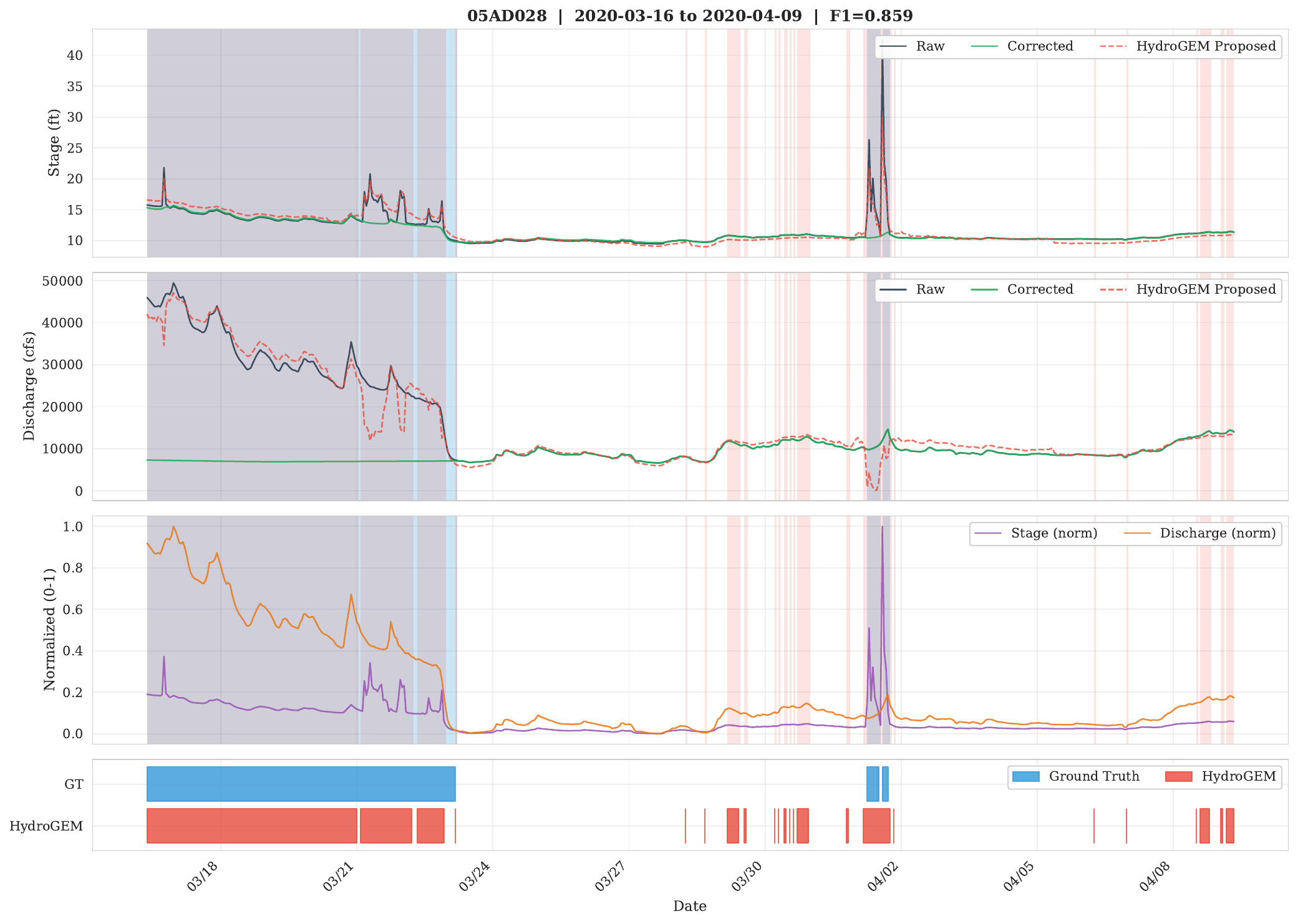}
    \caption{Detection example for Canadian station 05AD028 (F1 = 0.859). Blue and red shading indicate periods where ground truth and HydroGEM identify anomalies, respectively. The model preserves clean observations (late March) while flagging data quality issues. Reconstruction divergence during ice-affected periods (early March) reflects differences between USGS training corrections and ECCC operational practices~\cite{Turcotte2013,Beltaos2013}.}
    \label{fig:eccc_example}
\end{figure}

Figure~\ref{fig:eccc_example} illustrates HydroGEM's detection behavior on station 05AD028 during a spring period encompassing ice-affected and open-water conditions. The station achieved F1 = 0.859, demonstrating strong agreement with operational corrections. During clean periods (late March), the model correctly preserves data integrity with minimal false positives. The major stage spike on April 2 is appropriately flagged and reconstructed.

The ice-affected period (March 16--24, blue shading) exhibits reconstruction divergence where ECCC operators applied aggressive discharge reductions while stage remained elevated, a characteristic ice-backwater correction strategy. HydroGEM's proposed reconstructions reflect discharge-stage relationships learned from USGS training data, which differ systematically from Canadian practices in ice-control periods. This divergence does not indicate detection failure; rather, it demonstrates that the model learned USGS correction principles rather than memorizing site-specific patterns. For Canadian deployment, reconstructions should be treated as suggestions requiring operator review, particularly during ice-affected periods.

\subsubsection{Evaluation Under Weak Labels}
\label{subsubsec:canadian_multimetric}

The ECCC ground-truth labels are derived from operational corrections applied by Water Survey of Canada technicians. These corrections reflect practical data quality decisions rather than precise anomaly boundaries: a correction recorded as starting at midnight may correspond to an anomaly whose true onset occurred hours earlier. Standard pointwise F1 penalizes detections that correctly identify anomaly events but with minor temporal offset, a known limitation for time series anomaly detection on weakly labeled data~\cite{kim2022rigorous}.

To address this, we adopt evaluation metrics from the time series anomaly detection literature that account for temporal tolerance and interval-level detection. Tolerant F1 credits a prediction if it falls within $\pm$24 hours of a labeled correction timestep, accommodating the daily granularity typical of correction records~\cite{sorbo2024navigating}. Segment-level recall treats contiguous anomaly intervals as discrete events rather than independent timestamps and counts an event as detected if any predicted anomaly overlaps the interval or falls within the tolerance window~\cite{hundman2018detecting}.

Table~\ref{tab:canadian_multimetric} presents HydroGEM's performance across these complementary metrics.

\begin{table}[htbp]
\centering
\caption{Zero-shot HydroGEM performance on 100 ECCC stations (1\% correction threshold).}
\label{tab:canadian_multimetric}
\small
\begin{tabular}{@{}lccc@{}}
\toprule
Metric & Precision & Recall & F1/Score \\
\midrule
Pointwise F1 & 0.650 & 0.567 & 0.582 \\
Tolerant F1 ($\pm$24h) & 0.683 & 0.765 & \textbf{0.700} \\
Segment-level recall & --- & \textbf{0.901} & --- \\
\bottomrule
\end{tabular}
\end{table}

\begin{figure}[htbp]
    \centering
    \includegraphics[width=\columnwidth]{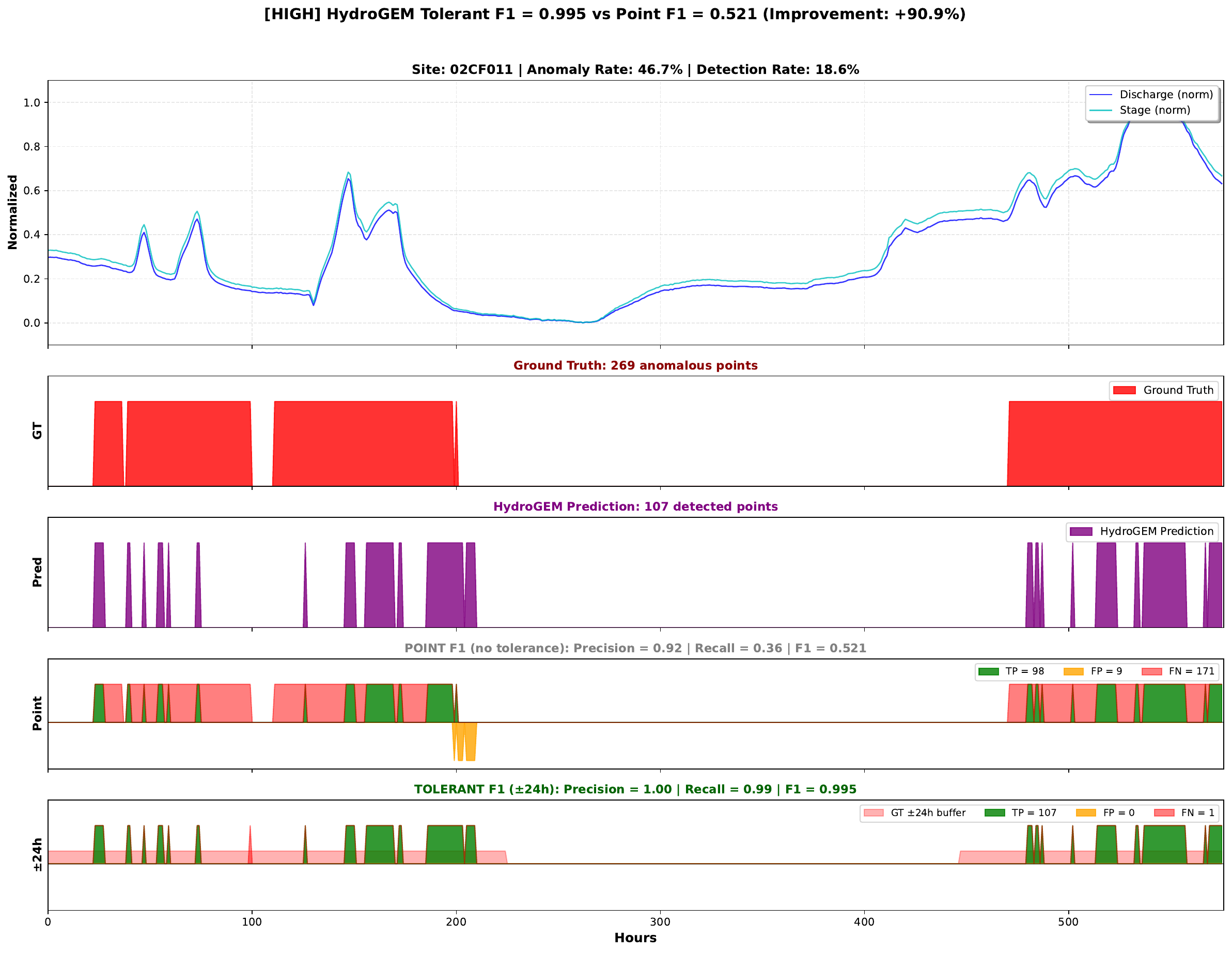}
    \caption{Illustration of tolerant evaluation on Canadian station 02CF011. Pointwise F1 (0.521) penalizes boundary misalignment despite correct event detection; tolerant F1 with $\pm$24-hour buffer (0.995) credits temporally proximate predictions.}
    \label{fig:tolerance_example}
\end{figure}

We adopt Tolerant F1 = 0.70 as the primary metric for this evaluation based on three considerations. First, tolerant evaluation aligns with operational practice: technicians review flagged periods holistically rather than verifying individual timestamps, so detections within 24 hours of recorded corrections remain operationally valuable. Second, the $\pm$24-hour buffer accommodates inherent timing imprecision in correction records without being excessively permissive. Third, precision remains stable across all buffer sizes tested (0.654 at $\pm$1h to 0.683 at $\pm$24h), indicating that the tolerance does not artificially inflate performance but simply avoids penalizing temporally proximate detections.

At the segment level, HydroGEM detects 90.1\% of anomaly events with at least partial overlap (segment recall = 0.901), demonstrating effective identification of anomaly intervals rather than just individual points. Extended evaluation including tolerance sensitivity analysis, weighted F1, and range score is provided in Appendix~\ref{app:eccc_metrics}.

\subsubsection{Detection Behavior Characterization}
\label{subsubsec:canadian_behavior}

\begin{figure}[htbp]
    \centering
    \includegraphics[width=\textwidth]{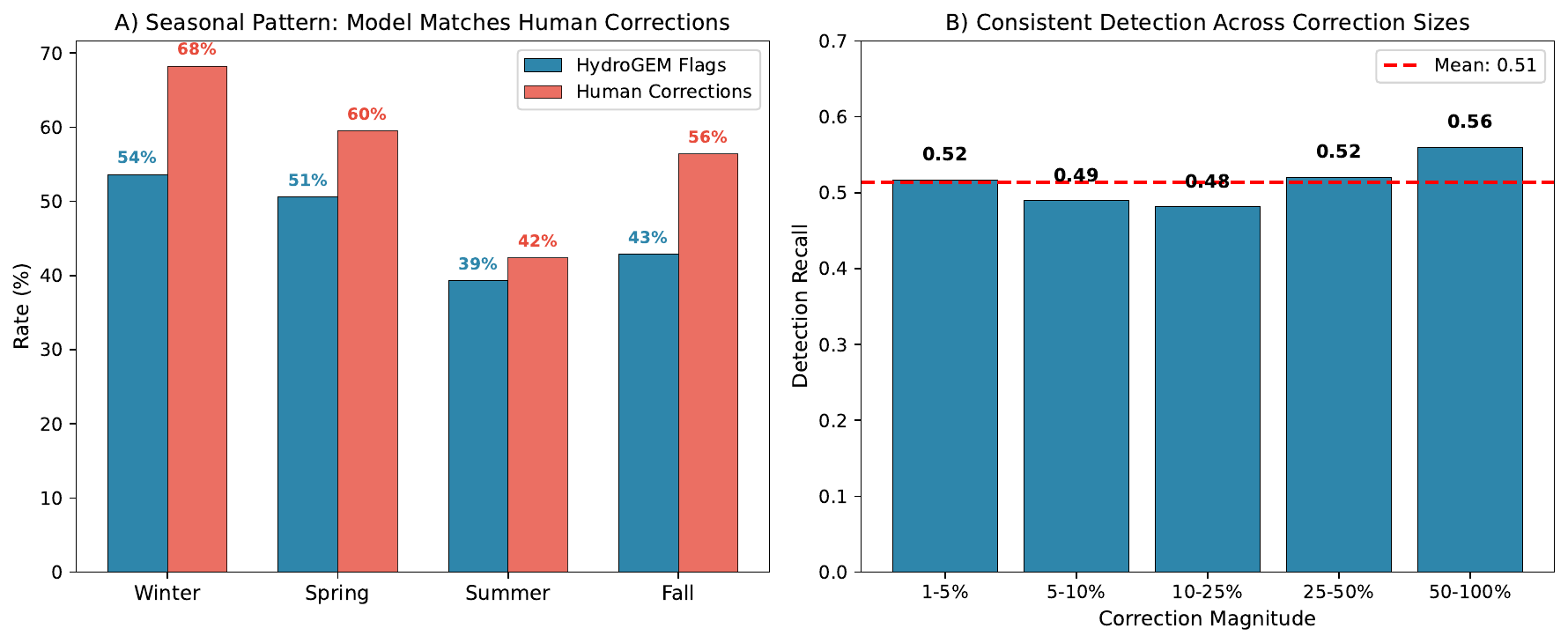}
    \caption{Detection behavior on Canadian zero-shot evaluation. (A) Seasonal comparison of HydroGEM flag rate versus human correction rate, showing aligned patterns with winter peaks. (B) Detection recall by correction magnitude, demonstrating consistent performance across small to large corrections.}
    \label{fig:detection_behavior}
\end{figure}

Beyond aggregate metrics, we analyzed whether HydroGEM learned physically meaningful patterns that transfer to Canadian conditions. Figure~\ref{fig:detection_behavior} summarizes detection behavior across seasons and correction magnitudes.

\paragraph{Seasonal alignment.}
We compared HydroGEM's flag rate against the human correction rate across seasons (Figure~\ref{fig:detection_behavior}A). Both exhibit consistent seasonal structure: winter shows the highest activity (HydroGEM: 54\%, human: 68\%), followed by spring (51\%, 60\%), fall (43\%, 56\%), and summer (39\%, 42\%). The winter peak reflects well-documented challenges in cold-region hydrometry: ice-affected flow produces backwater conditions violating standard rating curves, anchor ice and frazil ice cause sensor interference, and instrument freezing creates data gaps~\cite{WMO2010a,Turcotte2013,Beltaos2013}. That HydroGEM, trained on predominantly ice-free USGS stations, identifies these signatures in Canadian data suggests the model learned generalizable representations of sensor malfunction and rating curve violation rather than site-specific patterns.

\paragraph{Magnitude-independent detection.}
A concern with learned detectors is whether they simply flag large deviations while missing subtle corrections. Figure~\ref{fig:detection_behavior}B stratifies detection recall by correction magnitude. HydroGEM maintains consistent recall (0.48--0.56) across corrections ranging from minor adjustments (1--5\%) to major revisions (50--100\%), indicating the model learned anomaly patterns based on temporal dynamics, channel inconsistencies, and physical constraint violations rather than relying on deviation magnitude as a proxy.

\subsection{Ablation Studies}
\label{subsec:ablation}

We conducted systematic ablation studies to validate key design decisions. These experiments, performed on validation subsets during development, reveal the necessity of each architectural component for achieving strong zero-shot generalization.

\subsubsection{Normalization Strategy Impact}
\label{subsubsec:norm_ablation}

\begin{figure}[htbp]
\centering
\includegraphics[width=\columnwidth]{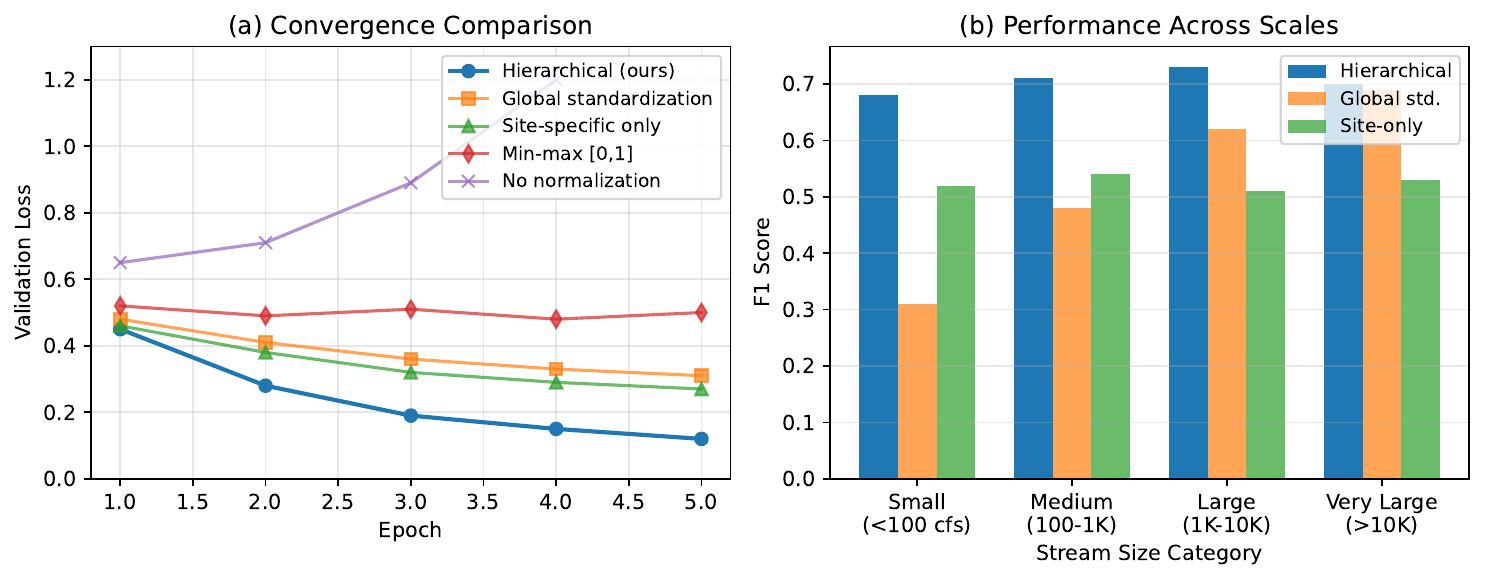}
\caption{Training dynamics under different normalization strategies. (a) Convergence comparison showing only hierarchical normalization achieves stable optimization. (b) Performance across stream size categories reveals severe bias in global standardization toward large rivers.}
\label{fig:normalization_convergence}
\end{figure}

The hierarchical normalization scheme emerged from systematic failures when applying standard approaches to continental-scale data. Figure~\ref{fig:normalization_convergence} shows convergence behavior across different normalization strategies on a 100-site validation subset.

Raw units cause gradient explosion as discharge values spanning six orders of magnitude create unstable optimization. Global standardization converges but exhibits severe bias: small streams achieve F1 = 0.31 while large rivers reach F1 = 0.69, indicating the model essentially ignores small-scale hydrology. Site-specific normalization alone prevents cross-site learning, achieving mediocre uniform performance (F1 $\approx$ 0.52) across all scales. Min-max normalization fails to converge as single outlier events repeatedly reshape the normalization range.

Our hierarchical approach (log transform $\rightarrow$ site standardization $\rightarrow$ global clipping with scale embeddings) represents the minimal configuration achieving three requirements simultaneously: stable gradients across training, meaningful cross-site comparison, and preservation of absolute scale information through embeddings. The scale embeddings ($\sigma_{\ln Q}$, $\sigma_{\ln H}$) prove particularly critical, reintroducing magnitude information that enables scale-appropriate anomaly detection thresholds.

\subsubsection{Architectural Component Analysis}
\label{subsubsec:arch_ablation}

\begin{figure}[htbp]
\centering
\includegraphics[width=\columnwidth]{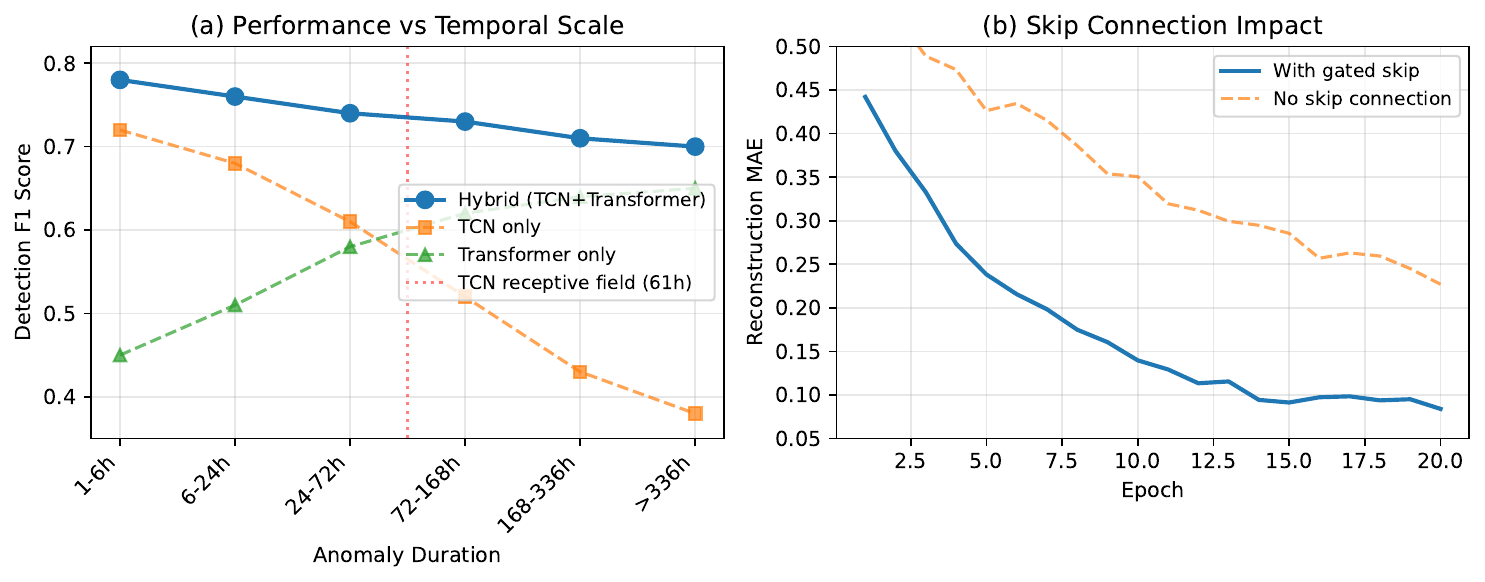}
\caption{Architectural ablation results. (a) Detection performance by anomaly duration shows TCN-only degradation beyond its 61-hour receptive field, while Transformer-only fails on short events. (b) Training curves demonstrate 2$\times$ slower convergence without gated skip connections.}
\label{fig:architecture_ablation}
\end{figure}

The hybrid TCN-Transformer architecture addresses fundamental limitations of using either component in isolation. Figure~\ref{fig:architecture_ablation} illustrates performance degradation at different temporal scales when architectural components are removed.

The TCN-only variant maintains strong performance (F1 $>$ 0.70) for anomalies under 24 hours but degrades precipitously for extended events. With receptive field mathematically limited to 61 hours (4 blocks, maximum dilation 8), the model cannot capture multi-week sensor drift or seasonal rating changes. Performance drops below F1 = 0.40 for anomalies exceeding 336 hours.

Conversely, the Transformer-only variant shows opposite behavior: poor performance on short transients (F1 = 0.45 for 1--6 hour events) but improving with duration. Self-attention struggles to isolate brief spikes within 576-hour windows, effectively smoothing high-frequency patterns that constitute many operational anomalies.

The gated skip connection facilitates stable early training. Without it, convergence requires approximately 20 epochs versus 10 with the skip path, and final reconstruction error remains 35\% higher. The gate parameter $\alpha$ is learned during training, starting near zero and increasing as the model converges, progressively incorporating learned representations while maintaining gradient flow.

\subsubsection{Training Strategy Validation}
\label{subsubsec:training_ablation}

Table~\ref{tab:training_ablation} quantifies the benefit of two-stage training through controlled experiments on 50-site validation subsets. Each variant used identical architecture and hyperparameters, differing only in training methodology.

\begin{table}[htbp]
\centering
\caption{Training strategy comparison on validation subset. Metrics averaged across three random seeds with standard deviation.}
\label{tab:training_ablation}
\small
\begin{tabular}{@{}lccc@{}}
\toprule
\textbf{Training Strategy} & \textbf{Detection F1} & \textbf{Clean Preservation} & \textbf{Convergence} \\
\midrule
\textbf{Two-stage (Ours)} & \textbf{0.684 $\pm$ 0.021} & \textbf{96.8\%} & 12 epochs \\
End-to-end synthetic & 0.521 $\pm$ 0.034 & 89.3\% & 8 epochs \\
No pretraining & 0.398 $\pm$ 0.048 & 91.2\% & 15+ epochs \\
Pretrain only (no Stage 2) & 0.216 $\pm$ 0.019 & 98.1\% & N/A \\
\bottomrule
\end{tabular}
\end{table}

End-to-end training with synthetic anomalies from random initialization shows 31\% lower F1 and higher variance, suggesting overfitting to specific corruption patterns. The model learns to detect training anomalies but fails to generalize to novel corruption types. Training without pretraining yields poorest detection performance and highest variance, as the model lacks robust priors about normal hydrological behavior.

The pretrained backbone provides stable initialization with learned representations of recession curves, rating relationships, and seasonal patterns. This foundation enables rapid Stage 2 convergence and superior final performance while maintaining 96.8\% clean data preservation---critical for operational trust.

\subsubsection{Loss Component Importance}
\label{subsubsec:loss_ablation}

\begin{figure}[htbp]
\centering
\includegraphics[width=0.8\columnwidth]{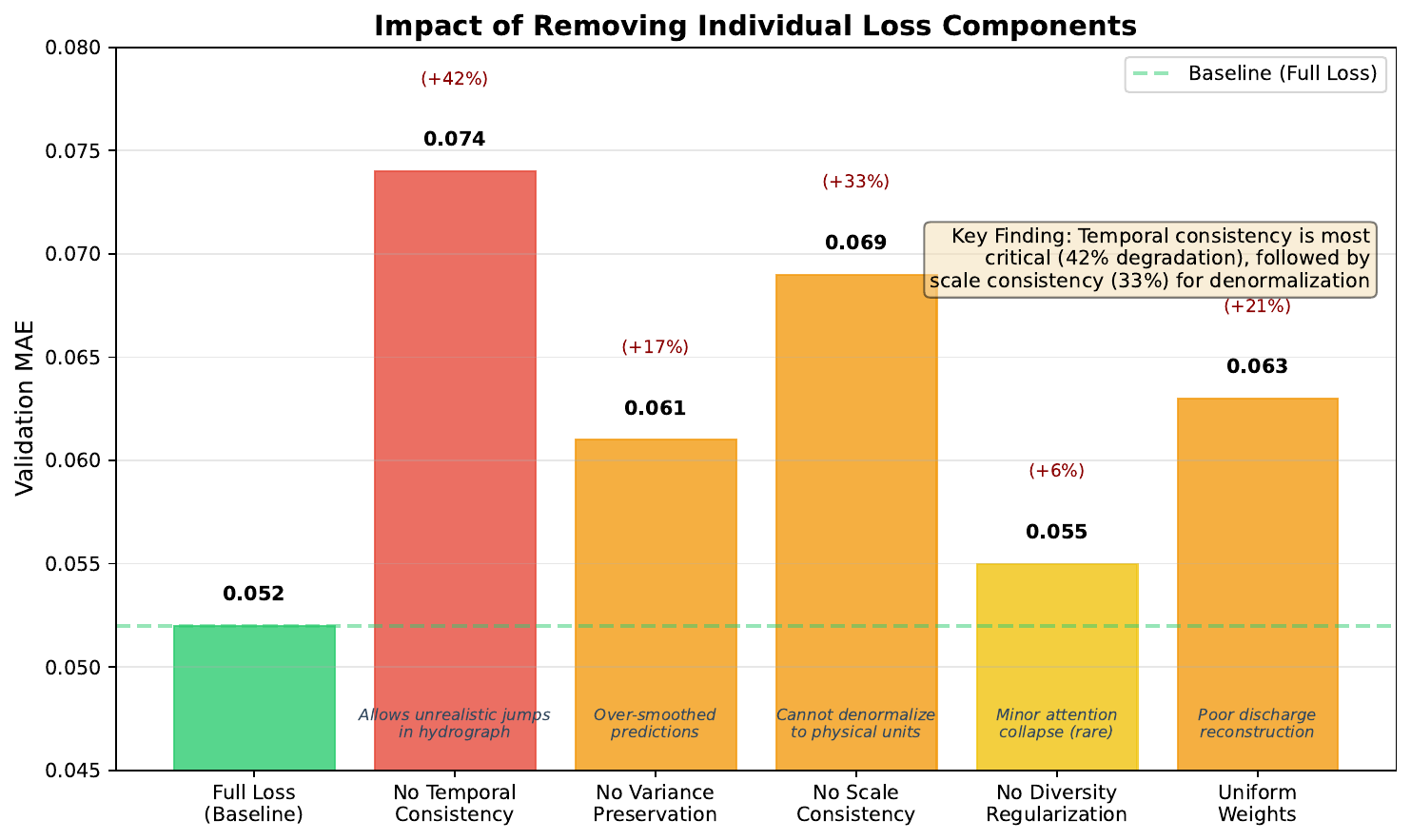}
\caption{Loss component ablation ordered by impact. Removing temporal consistency causes 42\% degradation, while scale consistency loss increases error by 33\%. Each component addresses operational requirements.}
\label{fig:loss_ablation}
\end{figure}

Each pretraining loss component prevents specific failure modes. Figure~\ref{fig:loss_ablation} shows validation error when individual components are removed, ordered by impact severity.

Temporal consistency loss proves most critical, with removal increasing MAE by 42\%. Without this constraint, reconstructions exhibit physically implausible discontinuities---sudden jumps in discharge while stage remains constant---that would trigger numerous false positives during detection. Scale consistency loss (33\% degradation) enables accurate denormalization to physical units, essential for providing meaningful corrections to operators. Variance preservation prevents over-smoothed predictions that would miss legitimate flow variability.

The weighted reconstruction loss focusing on discharge (weight = 3.0) and stage (weight = 2.5) over static features reflects operational priorities. Uniform weighting degrades performance by 21\%, particularly for discharge reconstruction where accuracy is paramount for water resource management decisions.

\subsubsection{Synthesis of Ablation Findings}
\label{subsubsec:ablation_synthesis}

These ablations demonstrate that HydroGEM's performance emerges from synergistic interaction between components rather than any single innovation. Hierarchical normalization allows learning across six orders of magnitude while preserving scale-aware detection capabilities. The hybrid architecture captures complementary temporal scales, with TCN handling short transients and Transformer modeling extended dependencies. Two-stage training prevents overfitting to synthetic patterns while establishing robust hydrological priors. The multi-component loss ensures physically plausible reconstructions meeting operational requirements.

Performance degrades 20--42\% when removing major components, confirming the current architecture represents a minimal configuration for continental-scale deployment. Each design choice addresses specific failure modes encountered when scaling anomaly detection beyond individual sites to thousands of heterogeneous monitoring stations.

\section{Discussion}
\label{sec:discussion}

\subsection{Interpretation of Key Findings}
\label{subsec:interpretation}

HydroGEM demonstrates that foundation model principles can be effectively adapted to operational hydrological quality control, addressing a critical bottleneck in environmental monitoring infrastructure. The results reveal several important insights about both the model's capabilities and the nature of streamflow data quality challenges.

The two-stage training approach proves effective at learning generalizable hydrometric principles without extensive labeled anomaly datasets. By pretraining on 6.03 million clean sequences, HydroGEM develops robust representations of normal hydrological behavior that serve as an inductive bias for anomaly detection. Recession curves, rating relationships, and seasonal patterns emerge naturally from the self-supervised objective. The subsequent Stage 2 training with simplified synthetic anomalies forces the model to learn fundamental consistency principles rather than memorizing specific corruption signatures. The four-axis separation between training and test distributions---equation form, parameter range, channel coupling, and normalized versus physical space---provides evidence that the 2.0$\times$ F1 improvement over baselines reflects genuine abstraction rather than pattern matching.

The hierarchical normalization scheme successfully enables learning across six orders of magnitude in discharge, from ephemeral desert washes to major rivers, within a single model. This addresses a fundamental challenge in continental-scale hydrology where traditional approaches either require site-specific calibration or suffer from large-river dominance. The combination of logarithmic stabilization, site-specific standardization, and scale embeddings preserves both local patterns and absolute magnitude information. The model can thus distinguish anomalies in small creeks from those in major rivers, despite their similar appearance in normalized space. This capability is essential for continental-scale deployment where manual calibration of thousands of sites would be prohibitively expensive.

Zero-shot transfer to Canadian stations demonstrates that learned representations capture general principles of streamflow monitoring rather than USGS-specific artifacts. The alignment between HydroGEM's detection patterns and operational correction rates across seasons is particularly notable. The winter peak in both model flagging and human corrections, corresponding to documented ice effects~\cite{Turcotte2013,Beltaos2013}, emerges despite training predominantly on ice-free stations. This suggests the model learned to identify fundamental signatures of sensor malfunction and rating curve violation that manifest similarly across different hydrological contexts. The ability to generalize across national boundaries, instrumentation protocols, and climatic regimes supports the foundation model approach for operational hydrology.

\subsection{Limitations and Operational Considerations}
\label{subsec:limitations}

Despite strong performance, several limitations require careful consideration for operational deployment. These constraints arise from both technical architecture decisions and fundamental challenges in defining ground truth for hydrological data quality.

Reconstruction divergence in ice-affected periods represents the most significant operational challenge. While HydroGEM successfully detects ice-related anomalies with high recall, reconstructions during these periods often diverge from operational corrections, particularly for Canadian stations. This divergence reflects fundamental differences in correction philosophy between agencies. USGS guidance emphasizes maintaining physical plausibility of both discharge and stage, typically applying moderate adjustments that preserve the coupled relationship. In contrast, ECCC practice often reduces discharge substantially while preserving elevated stage readings during ice backwater, acknowledging that stage may accurately represent local conditions even when unsuitable for discharge computation~\cite{eccc_ice_warning}. This is not a detection failure but rather highlights that appropriate corrections depend on operational context, downstream use cases, and agency-specific protocols. For deployment, we recommend treating all reconstructions as suggestions requiring review, particularly during known ice periods.

The model occasionally flags rapid but valid discharge fluctuations from gate operations, hydropeaking, or flash floods as anomalous. These events produce sudden excursions in discharge-stage relationships that resemble sensor artifacts in the learned representation. While maintaining high recall ensures genuine anomalies are rarely missed, this behavior reinforces the necessity of human oversight. The challenge of distinguishing anthropogenic flow modifications from sensor failures without explicit metadata represents a fundamental limitation of approaches based solely on time series patterns.

Training requires substantial computational resources, with backbone pretraining consuming approximately 48 GPU-hours on modern hardware. While this one-time cost amortizes across thousands of deployment sites, it may limit accessibility for smaller agencies or research groups. Inference remains efficient at less than 100ms per window on standard hardware, enabling real-time deployment. However, the current architecture processes 576-hour windows in batch mode, creating a tradeoff between temporal context and streaming responsiveness.

The model operates solely on observable quantities without meteorological forcing, maximizing applicability to gauge-only stations but preventing discrimination between sensor failures and physical extremes using precipitation context. A large discharge spike could represent sensor malfunction or an actual flash flood; without rainfall data, the model cannot definitively distinguish these cases.

\subsection{Future Directions}
\label{subsec:future}

Several extensions could address current limitations and expand applicability. Integration of multivariate causal information represents a natural next step. Incorporating precipitation time series, upstream gauge observations, and reservoir release schedules would enable the model to reason about physical causes of observed patterns. For example, a discharge spike coinciding with heavy rainfall across the watershed likely reflects genuine hydrological response rather than sensor malfunction, while an isolated spike without precipitation or upstream response suggests instrumentation issues. Such causal reasoning could substantially reduce false positives on legitimate extreme events.

Development of streaming architectures would enable real-time deployment with incremental updates rather than batch processing. Architectures supporting online inference while maintaining long-term memory could reduce detection latency from hours to minutes, better matching operational monitoring requirements.

Integration with large language models offers potential for automated report generation. Detected anomalies could be summarized in natural language explanations describing the likely failure mode, affected time periods, and recommended actions, reducing cognitive load on operators reviewing flagged observations.

Extension to water quality parameters presents a natural application domain where sensor drift, biofouling, and calibration issues produce similar quality control challenges. The two-stage training framework and hierarchical normalization could transfer to dissolved oxygen, turbidity, conductivity, and other continuously monitored water quality variables.

Finally, active learning frameworks that continuously improve from operator correction feedback could enable ongoing model refinement as deployment generates new labeled examples, progressively improving discrimination in challenging cases like ice effects and controlled releases.

\section{Conclusion}
\label{sec:conclusion}

This work introduced HydroGEM, a foundation model for continental-scale streamflow quality control that addresses the growing mismatch between automated data generation and review capacity in hydrological monitoring networks. Through careful architectural design and training strategies, we demonstrated that foundation model principles can be successfully adapted to operational sensor networks despite challenges including extreme heterogeneity, sparse labeling, and stringent deployment constraints.

The technical contributions center on three innovations. First, the two-stage training approach combines self-supervised pretraining on 6.03 million sequences with Stage 2 training using synthetic anomalies, eliminating dependence on scarce labeled datasets while learning robust representations of hydrological normality. Second, the hierarchical normalization scheme enables learning across six orders of magnitude within a single model, addressing a key challenge in continental-scale hydrology. Third, the rigorous evaluation framework with synthetic anomalies and zero-shot cross-national transfer provides evidence for genuine generalization rather than pattern memorization.

Experimental validation confirms the effectiveness of these approaches. On synthetic test data with known ground truth, HydroGEM achieves F1 = 0.792 with 68.7\% reconstruction error reduction, approximately 2.0$\times$ the F1 of the strongest baseline while maintaining 96.8\% preservation of clean data. Zero-shot transfer to 100 Canadian stations yields pointwise F1 = 0.582 and tolerant F1 = 0.70, with detection patterns aligning with operational correction rates across seasons. The model identifies all 18 anomaly types and maintains consistent performance across correction magnitudes from 1\% to 100\%, indicating pattern-based rather than threshold-based detection.

As environmental monitoring transitions from sparse manual sampling to dense automated sensing, quality control emerges as the critical bottleneck preventing full utilization of these data streams for scientific understanding and operational decision-making. HydroGEM demonstrates that foundation models can help bridge this gap through human-AI collaboration that preserves oversight while enabling continental-scale monitoring. This approach, where AI handles pattern recognition at scale while humans provide domain expertise and accountability, may prove essential as we build the observational infrastructure needed to understand and manage water resources under accelerating environmental change.

\section*{Declaration of Generative AI and AI-Assisted Technologies in the Writing Process}

During the preparation of this manuscript, the authors used ChatGPT (OpenAI) to assist with grammar and language refinement in the Introduction and Discussion sections. After using this tool, the authors reviewed and edited all content as needed and take full responsibility for the content of the publication.


\section*{Acknowledgments}

This material is based upon work supported by the National Science Foundation under Grant No. EAR 2012123. Any opinions, findings, and conclusions or recommendations expressed in this material are those of the author(s) and do not necessarily reflect the views of the National Science Foundation. Any use of trade, firm, or product names is for descriptive purposes only and does not imply endorsement by the U.S. Government.

The work was also supported by the University of Vermont College of Engineering and Mathematical Sciences through the REU program.

The authors acknowledge the U.S. Geological Survey (USGS) for providing access to the National Water Information System and for domain expertise that informed the anomaly taxonomy. We also thank Environment and Climate Change Canada (ECCC) for providing access to hydrometric data through the Water Survey of Canada.

The authors acknowledge the National Artificial Intelligence Research Resource (NAIRR) Pilot and the Texas Advanced Computing Center (TACC) at The University of Texas at Austin for providing computational resources under Award No. NAIRR240491.

\section*{Declaration of Competing Interest}

The authors declare that they have no known competing financial interests or personal relationships that could have appeared to influence the work reported in this paper.

\section*{Data Availability}

USGS streamflow observations are publicly available through the National Water Information System (NWIS) at \url{https://waterdata.usgs.gov/nwis}. Environment and Climate Change Canada hydrometric data are available through the Water Survey of Canada at \url{https://wateroffice.ec.gc.ca/}. Model weights, inference code, representative detection examples, and documentation describing the synthetic benchmark construction and Canadian station selection are available at \url{https://huggingface.co/Ejokhan/HydroGEM}.

\section*{CRediT Author Statement}

\textbf{Ijaz Ul Haq}: Conceptualization, Methodology, Software, Validation, Formal analysis, Investigation, Data curation, Writing - Original Draft, Visualization.
\textbf{Byung Suk Lee}: Supervision, Resources, Writing - Review \& Editing, Funding acquisition.
\textbf{Julia N. Perdrial}: Supervision, Domain expertise, Review \& Editing, Funding acquisition.
\textbf{David Baude}: Data curation, Writing - Review \& Editing.


\appendix

\section{Hierarchical Normalization}
\label{app:normalization}

This appendix provides complete mathematical derivations and implementation details for the three-tier hierarchical normalization scheme introduced in Section~\ref{subsubsec:normalization}.

\subsubsection{Tier 1: Logarithmic Stabilization}

For approximately log-normal hydrological variables~\cite{sangal1970lognormal}, we apply:

\begin{equation}
Q^{(1)} = \ln(Q + \epsilon), \quad H^{(1)} = \ln(H + \epsilon)
\end{equation}

\noindent with $\epsilon = 10^{-8}$ to handle near-zero flows in ephemeral systems.

\textbf{Rationale}: This transformation achieves three critical objectives:

\paragraph{1. Rating curve linearization.} Natural channel rating curves follow power-law form $Q = a(H - H_0)^b$~\cite{chow1988applied}. In log-space:

\begin{equation}
\ln Q \approx b \ln H + \ln a
\end{equation}

This linearization enables the network to learn discharge-stage coupling through simple linear relationships rather than complex nonlinear functions.

\paragraph{2. Multiplicative-to-additive noise conversion.} Hydrological measurement errors are often proportional to magnitude (e.g., 5\% uncertainty applies to both 10 ft$^3$/s and 10,000 ft$^3$/s). In linear space, this multiplicative noise has magnitude-dependent variance:

\begin{equation}
\text{Var}(Q_{\text{measured}}) = (0.05 \cdot Q_{\text{true}})^2
\end{equation}

The log transformation converts multiplicative noise to additive noise with constant variance:

\begin{equation}
\text{Var}(\ln Q_{\text{measured}}) \approx \text{const}
\end{equation}

\paragraph{3. Heteroscedasticity stabilization.} Without log transformation, gradient magnitudes for a 10\% error on a large river exceed those for small streams by $10^3$-$10^6\times$, preventing convergence. The log transformation ensures equal-percentage errors produce equal-magnitude gradients regardless of absolute scale.

\subsubsection{Tier 2: Site-Specific Standardization}

For each site $s$, we compute normalization statistics exclusively from that site's training partition:

\begin{equation}
\mu_s = \mathbb{E}_{t \in \text{train}(s)}[\mathbf{x}_t^{(1)}], \quad \sigma_s = \text{Std}_{t \in \text{train}(s)}[\mathbf{x}_t^{(1)}]
\end{equation}

\noindent and transform:

\begin{equation}
\mathbf{x}_t^{(2)} = \frac{\mathbf{x}_t^{(1)} - \mu_s}{\sigma_s + \epsilon}
\end{equation}

\noindent with $\epsilon = 10^{-8}$ for numerical stability.

\textbf{Critical implementation detail - preventing data leakage}: For sites appearing only in validation or test partitions, we substitute \emph{global training statistics} computed over all training sites:

\begin{equation}
\mu_{\text{global}} = \mathbb{E}_{(s,t)\in \mathcal{D}_{\text{train}}}[\mathbf{x}_{t}^{(1)}], \quad
\sigma_{\text{global}} = \text{Std}_{(s,t)\in \mathcal{D}_{\text{train}}}[\mathbf{x}_{t}^{(1)}]
\end{equation}
where $\mathcal{D}_{\text{train}}$ is the union of all timesteps across all training sites.

This ensures \emph{strict partition separation}---validation and test data never influences normalization parameters. Most multi-site hydrological models violate this principle by computing statistics over all sites (including validation/test), creating subtle information leakage.

\textbf{Rationale}: Site-specific standardization enables weight sharing across basins by placing each site's typical range in a common normalized space, preventing large rivers from dominating the loss landscape. A small stream with $Q \in [1, 10]$ ft$^3$/s and the Mississippi River with $Q \in [10^4, 10^6]$ ft$^3$/s both normalize to $\approx[-2, +2]$, contributing equally to gradient updates.

\subsubsection{Tier 3: Global Clipping}

To downweight extreme leverage points (measurement spikes, brief sensor failures, telemetry errors), we clip standardized variables to a symmetric bound:

\begin{equation}
\mathbf{x}_t^{\text{norm}} = \text{clip}(\mathbf{x}_t^{(2)}, -\tau, +\tau), \quad \tau = 3.0
\end{equation}

\textbf{Critical asymmetry}: Clipping applies \emph{only to inputs} during training. Predicted outputs are denormalized \emph{without clipping} to avoid artificially truncating physical predictions.

\textbf{Rationale}: For normally distributed standardized variables, $\tau = 3.0$ retains approximately 99.7\% of values while capping extreme outliers. In long 576-hour sequences, even rare extreme values can cause gradient explosion. Clipping provides numerical stability while preserving the network's ability to reconstruct full-range physical magnitudes through the exact inverse mapping.

\subsubsection{Exact Inverse Mapping}

Let $\hat{y}^{\text{norm}} \in \mathbb{R}^{12}$ denote a model output on the normalized scale. Recovery to physical units follows the reverse transformation chain:

\begin{align}
\hat{y}^{(1)} &= \hat{y}^{\text{norm}} \cdot (\sigma_s + \epsilon) + \mu_s \label{eq:inv_standardize} \\
\hat{y} &= \exp(\hat{y}^{(1)}) - \epsilon \label{eq:inv_log}
\end{align}

\noindent where Eq.~\ref{eq:inv_standardize} reverses standardization and Eq.~\ref{eq:inv_log} reverses log transformation.

\textbf{Special cases}:

\begin{itemize}
\item If site $s$ used global statistics during normalization (validation/test site), apply the same global statistics in the inverse
\item Static variables never log-transformed (latitude, longitude, elevation, drainage area) skip Eq.~\ref{eq:inv_log}
\end{itemize}

\textbf{Numerical verification}: We verify exact invertibility by round-trip transformation:

\begin{equation}
\|\mathbf{x}_{\text{original}} - \text{denorm}(\text{norm}(\mathbf{x}_{\text{original}}))\|_{\infty} < 10^{-6}
\end{equation}

\noindent for all training, validation, and test sequences.

\subsubsection{Scale Embeddings}

The scale embeddings $\{\sigma_{\ln Q,s}, \sigma_{\ln H,s}\}$ complete the normalization system by returning information about absolute variability lost during standardization:

\begin{equation}
\sigma_{\ln Q,s} = \text{Std}_{t \in \text{train}(s)}[\ln(Q_t + \epsilon)], \quad \sigma_{\ln H,s} = \text{Std}_{t \in \text{train}(s)}[\ln(H_t + \epsilon)]
\end{equation}

These embeddings are broadcast across the temporal dimension and concatenated with the normalized features $\mathbf{x}^{\text{norm}}$, enabling the network to condition on absolute scale.

\textbf{Rationale}: After site-specific standardization, a normalized discharge value of $+1.0$ is ambiguous---it could represent 10 ft$^3$/s at a small headwater stream or 10,000 ft$^3$/s at a large river. Both appear identical in normalized space. The scale embeddings resolve this ambiguity, enabling the network to express scale-dependent behavior (e.g., flashiness in small basins, baseflow dominance in large rivers) despite operating in normalized space.

\textbf{Design choice validation}: Ablation studies (Section~\ref{subsec:ablation}) confirm that removing scale embeddings substantially degrades detection performance, demonstrating their necessity for continental-scale learning.


\section{Architecture Equations}
\label{app:architecture}

This appendix provides complete mathematical formulations for the TCN-Transformer backbone described in Section~\ref{subsubsec:backbone}.

\subsection{TCN Encoder}

The dilated convolution operation for sequence $\mathbf{x}$ with filter $\mathbf{f}$ and dilation rate $r$:
\begin{equation}
(\mathbf{x} *_r \mathbf{f})(t) = \sum_{i=0}^{k-1} f(i) \cdot x_{t-r \cdot i}
\end{equation}

Linear projection from input to hidden dimension:
\begin{equation}
\mathbf{H}^{(0)} = \mathbf{X}\mathbf{W}_{\text{proj}} + \mathbf{b}_{\text{proj}}, \quad \mathbf{W}_{\text{proj}} \in \mathbb{R}^{12 \times 128}
\end{equation}

TCN block with residual connection:
\begin{equation}
\text{TCN-Block}(\mathbf{H}, r) = \mathbf{H} + \mathcal{F}(\mathbf{H}, r)
\end{equation}
where $\mathcal{F}(\mathbf{H}, r) = \text{Conv1D}_r(\text{ReLU}(\text{BN}(\text{Conv1D}_r(\mathbf{H}))))$.

Receptive field at block $\ell$:
\begin{equation}
\text{RF}_{\ell} = 2^{\ell+2} - 3
\end{equation}

\subsection{Transformer}

Projection to transformer dimension:
\begin{equation}
\mathbf{Z}^{(0)} = \mathbf{H}^{(4)}\mathbf{W}_{\text{up}} + \mathbf{b}_{\text{up}}, \quad \mathbf{W}_{\text{up}} \in \mathbb{R}^{128 \times 256}
\end{equation}

Query, key, value projections for head $h$:
\begin{equation}
\mathbf{Q}_h = \mathbf{Z}\mathbf{W}_{Q,h}, \quad \mathbf{K}_h = \mathbf{Z}\mathbf{W}_{K,h}, \quad \mathbf{V}_h = \mathbf{Z}\mathbf{W}_{V,h}
\end{equation}

L2 normalization for cosine similarity:
\begin{equation}
\hat{\mathbf{q}}_i = \frac{\mathbf{q}_i}{\|\mathbf{q}_i\|_2 + \epsilon}, \quad \hat{\mathbf{k}}_j = \frac{\mathbf{k}_j}{\|\mathbf{k}_j\|_2 + \epsilon}
\end{equation}

Attention scores with temporal decay and causal mask:
\begin{equation}
\mathbf{A}_{ij} = \frac{\hat{\mathbf{q}}_i^T \hat{\mathbf{k}}_j}{\sqrt{d_h}} \cdot \gamma^{|i-j|} \cdot \mathbb{1}_{i \geq j}
\end{equation}

Multi-head attention:
\begin{equation}
\text{MultiHead}(\mathbf{Z}) = \text{Concat}(\text{head}_1, \ldots, \text{head}_8)\mathbf{W}_O
\end{equation}

Feed-forward network:
\begin{equation}
\text{FFN}(\mathbf{x}) = \text{GELU}(\mathbf{x}\mathbf{W}_1 + \mathbf{b}_1)\mathbf{W}_2 + \mathbf{b}_2
\end{equation}

Layer normalization:
\begin{equation}
\text{LayerNorm}(\mathbf{x}) = \frac{\mathbf{x} - \mu}{\sqrt{\sigma^2 + \epsilon}} \cdot \gamma + \beta
\end{equation}

\subsection{Decoder and Skip Connections}

Decoder projection:
\begin{equation}
\mathbf{G}^{(0)} = \mathbf{Z}^{(L)}\mathbf{W}_{\text{down}} + \mathbf{b}_{\text{down}}, \quad \mathbf{W}_{\text{down}} \in \mathbb{R}^{256 \times 128}
\end{equation}

Final output projection:
\begin{equation}
\hat{\mathbf{X}}_{\text{decoder}} = \mathbf{G}^{(4)}\mathbf{W}_{\text{out}} + \mathbf{b}_{\text{out}}, \quad \mathbf{W}_{\text{out}} \in \mathbb{R}^{128 \times 12}
\end{equation}

Gated skip connection:
\begin{equation}
\hat{\mathbf{X}} = (1 - \sigma(\alpha)) \cdot \hat{\mathbf{X}}_{\text{decoder}} + \sigma(\alpha) \cdot \text{Linear}(\mathbf{H}^{(4)}_{\text{encoder}})
\end{equation}


\section{Loss Functions}
\label{app:losses}

\subsection{Pretraining Loss}

Combined pretraining objective:
\begin{equation}
\mathcal{L}_{\text{pretrain}} = \mathcal{L}_{\text{recon}} + 0.6\mathcal{L}_{\text{temporal}} + 0.4\mathcal{L}_{\text{variance}} + 0.3\mathcal{L}_{\text{scale}} + 0.05\mathcal{L}_{\text{diversity}}
\end{equation}

Weighted reconstruction loss:
\begin{equation}
\mathcal{L}_{\text{recon}} = \frac{1}{T} \sum_{t=1}^{T} \sum_{f=1}^{d} w_f \cdot (\hat{x}_{t,f} - x_{t,f})^2
\end{equation}
with weights $w = 3.0$ (discharge), $2.5$ (stage), $1.5$ (seasonal), $1.0$ (static).

Temporal consistency loss:
\begin{equation}
\mathcal{L}_{\text{temporal}} = \frac{1}{T-1} \sum_{t=1}^{T-1} \sum_{f \in \{Q,H\}} w_f \cdot [(\hat{x}_{t+1,f} - \hat{x}_{t,f}) - (x_{t+1,f} - x_{t,f})]^2
\end{equation}

Variance preservation loss:
\begin{equation}
\mathcal{L}_{\text{variance}} = \sum_{f=1}^{d} [\text{Var}(\hat{\mathbf{x}}_f) - \text{Var}(\mathbf{x}_f)]^2
\end{equation}

Scale consistency loss:
\begin{equation}
\mathcal{L}_{\text{scale}} = \frac{1}{T} \sum_{t=1}^{T} \sum_{f \in \text{scale}} (\hat{x}_{t,f} - x_{t,f})^2
\end{equation}

Diversity regularization:
\begin{equation}
\mathcal{L}_{\text{diversity}} = -\text{Entropy}(\mathbf{A}) + \lambda_{\text{rank}} \max(0, 10 - \text{rank}(\mathbf{H}))
\end{equation}

\subsection{Detection Head Training Loss}

Focal loss for detection:
\begin{equation}
\mathcal{L}_{\text{focal}} = -\frac{1}{T}\sum_{t=1}^{T} \alpha_t (1-\hat{p}_t)^\gamma [y_t \log \hat{p}_t + (1-y_t)\log(1-\hat{p}_t)]
\end{equation}
with $\alpha = 0.25$, $\gamma = 2.0$.

Corruption reconstruction loss:
\begin{equation}
\mathcal{L}_{\text{corrupt}} = \frac{1}{|\mathcal{T}_{\text{anom}}|} \sum_{t \in \mathcal{T}_{\text{anom}}} \|\hat{\mathbf{X}}_t - \mathbf{X}_t^{(\text{clean})}\|_2^2
\end{equation}

Clean preservation loss:
\begin{equation}
\mathcal{L}_{\text{preserve}} = \frac{1}{|\mathcal{T}_{\text{clean}}|} \sum_{t \in \mathcal{T}_{\text{clean}}} \|\hat{\mathbf{X}}_t - \mathbf{X}_t^{(\text{obs})}\|_2^2
\end{equation}

Physics constraint loss:
\begin{equation}
\mathcal{L}_{\text{physics}} = \frac{1}{T-1}\sum_{t=1}^{T-1} \text{ReLU}(-\nabla_t Q \cdot \nabla_t H) + \lambda_{\text{RC}} \mathbb{E}[d_{\text{RC}}]
\end{equation}
where $d_{\text{RC}}$ is the absolute residual from a local power-law rating fit in log space, computed over a sliding window and averaged over timesteps.

Combined Stage 2 objective:
\begin{equation}
\mathcal{L}_{\text{stage2}} = 1.5 \mathcal{L}_{\text{focal}} + \lambda_c(\epsilon) \mathcal{L}_{\text{corrupt}} + \lambda_p(\epsilon) \mathcal{L}_{\text{preserve}} + 0.1 \mathcal{L}_{\text{physics}}
\end{equation}
with $\lambda_c(\epsilon) = 0.25 \cdot \min(1, \epsilon/3)$ and $\lambda_p(\epsilon) = 0.05 \cdot \min(1, \epsilon/3)$.


\section{Quality Control Protocols}
\label{app:qc}

This appendix details the multi-tier quality control protocol applied to ensure training data integrity (Section~\ref{subsubsec:qc_preprocessing}).

\subsubsection{Outlier Detection}

Values exceeding four standard deviations from site-specific monthly means are flagged for manual review:

\begin{equation}
\text{flag}(Q_t) = \mathbb{I}\left[|Q_t - \mu_{Q,m(t),s}| > 4\sigma_{Q,m(t),s}\right]
\end{equation}

\noindent where $m(t)$ denotes the month of timestep $t$, and $\mu_{Q,m,s}$, $\sigma_{Q,m,s}$ are computed from all observations in month $m$ for site $s$ within the training period.

\textbf{Rationale}: The adaptive monthly threshold accounts for natural hydrologic variability (e.g., higher variance during spring snowmelt) while identifying extreme leverage points. Fixed global thresholds would flag normal high flows in flashy basins while missing subtle anomalies in stable systems.

\subsubsection{Physical Plausibility Checks}

\paragraph{Temporal consistency.} We test rate-of-change using a site-specific threshold:

\begin{equation}
\frac{|Q_t - Q_{t-1}|}{Q_{t-1}} > \theta_{\text{site}}
\end{equation}

\noindent where $\theta_{\text{site}}$ is computed as the 99th percentile of observed fractional changes during the training period for that site. This adaptive threshold accommodates both flashy small basins and slowly-responding large systems.

\paragraph{Rating curve validation.} Stage-discharge pairs are validated against the power-law rating form $Q = a(H - H_0)^b$~\cite{chow1988applied}. We fit this relationship using robust regression (RANSAC with 1000 iterations, inlier threshold 15\%) on each site's training data, then flag pairs deviating beyond 2$\times$ the fitted residual standard deviation.

\paragraph{Range bounds.} All values are checked against physically plausible ranges:

\begin{itemize}
\item Discharge: $Q \in [0, Q_{\max}]$ where $Q_{\max} = \max(\text{training data}) \times 2.0$
\item Stage: $H \in [H_{\min} - 1.0, H_{\max} + 1.0]$ ft, allowing 1-foot exceedance beyond observed training range
\end{itemize}

\subsubsection{Gap Filling Strategy}

\paragraph{Short gaps (1-6 hours): Linear interpolation}

\begin{equation}
Q_t = Q_{t_0} + (Q_{t_1} - Q_{t_0}) \cdot \frac{t - t_0}{t_1 - t_0}
\end{equation}

\noindent where $t_0$ and $t_1$ are the last valid timestep before and first valid timestep after the gap.

\textbf{Rationale}: Linear interpolation is appropriate for slowly-varying baseflow conditions where discharge changes smoothly. For 1-6 hour gaps, this assumption typically holds.

\paragraph{Medium gaps (6-24 hours): Exponential recession model}

For gaps exceeding 6 hours but less than 24 hours, we apply the exponential recession relationship~\cite{vogel1992baseflow,tallaksen1995hydrograph}:

\begin{equation}
Q(t) = Q(t_0) \exp[-k(t - t_0)]
\end{equation}

The recession constant $k$ is estimated via:

\begin{equation}
k = -\frac{1}{\Delta t} \ln\left(\frac{Q(t_0 + \Delta t)}{Q(t_0)}\right)
\end{equation}

\noindent using a 48-hour window before the gap. If insufficient pre-gap data exists, we use the site-specific median recession constant computed from all valid recession segments in the training period.

\textbf{Rationale}: The exponential form respects fundamental watershed storage dynamics: $\frac{dS}{dt} = -kS$, where storage $S$ is linearly related to discharge. This is more physically realistic than linear interpolation for medium-duration gaps where recession behavior dominates.

\paragraph{Long gaps ($>$24 hours): Exclusion}

Gaps exceeding 24 hours are excluded entirely from sequence construction. No imputation is attempted to prevent long-range artifacts that could introduce spurious patterns.

\subsubsection{Gap Statistics}

The complete gap-filling protocol retains 94.7\% of potential 576-hour sequences:

\begin{itemize}
\item 89.3\% contain \emph{no} interpolation
\item 9.8\% contain 1-3 interpolated hours (linear)
\item 0.9\% contain 4-6 hours (mix of linear and recession)
\item 0\% contain $>$24 hours (sequences intersecting gaps $>$24h are dropped)
\end{itemize}

\textbf{Leakage prevention}: Any sequence window intersecting a flagged timestep (outlier, physical implausibility, or gap $>$24h) is excluded entirely, even if only a single hour is affected. This conservative approach eliminates potential information leakage from imputation.

\section{Training-Time Anomaly Injection}
\label{app:train_injector}

This appendix provides implementation details for the on-the-fly synthetic anomaly injection during training (Section~\ref{subsec:anomaly_injection}).

\subsubsection{Simplified Anomaly Types}

The training injector implements approximately 11 simplified corruption patterns applied in normalized log-space:

\begin{enumerate}
\item \textbf{Spike}: Add Gaussian noise impulses: $x'_t = x_t + \mathcal{N}(0, \alpha \sigma_x)$ at random positions, $\alpha \sim U(2, 5)$

\item \textbf{Drift}: Apply linear trend: $x'_t = x_t + \beta \cdot (t - t_{\text{start}})$ where $\beta \sim U(-0.01, +0.01)$

\item \textbf{Flatline}: Freeze values: $x'_t = x_{t_{\text{freeze}}}$ for all $t$ in segment

\item \textbf{Dropout}: Replace with near-zero: $x'_t = \epsilon$ where $\epsilon \sim U(10^{-6}, 10^{-4})$

\item \textbf{Saturation}: Clip to limits: $x'_t = \text{clip}(x_t, x_{\min} + 0.1, x_{\max} - 0.1)$

\item \textbf{Clock shift}: Temporal offset: $x'_t = x_{t + \Delta t}$ where $\Delta t \sim \{-3, -2, -1, +1, +2, +3\}$ hours

\item \textbf{Quantization}: Discretize: $x'_t = \text{round}(x_t / \Delta q) \cdot \Delta q$ where $\Delta q \sim U(0.05, 0.2)$

\item \textbf{Unit jump}: Abrupt bias: $x'_t = x_t + \gamma$ where $\gamma \sim U(-1.0, +1.0)$

\item \textbf{Temporal warp}: Stretch/compress: resample segment via $t' = t_{\text{start}} + (t - t_{\text{start}}) \cdot w$ where $w \sim U(0.8, 1.2)$

\item \textbf{Splice}: Replace segment with values from different time: $\mathbf{x}'_{t_1:t_2} = \mathbf{x}_{t_3:t_3+(t_2-t_1)}$

\item \textbf{Subtle drift}: Very gentle linear trend: $x'_t = x_t + \beta \cdot (t - t_{\text{start}})$ where $\beta \sim U(-0.002, +0.002)$
\end{enumerate}

These simple transformations contrast sharply with test-time physical-space injections using multi-variant equations (exponential/sigmoid/polynomial drift, hydraulic-coupled ice effects, etc.).

\subsubsection{Coverage Control Algorithm}

To achieve target coverage $c_{\text{target}}$ (e.g., 10\% for light tier, 20\% for moderate tier), the injector uses iterative refinement:

\begin{enumerate}
\item Initialize: $n_{\text{segments}} \sim U(2, 4)$, strength $\alpha = 1.0$
\item For attempt = 1 to 3:
\begin{enumerate}
\item Generate $n_{\text{segments}}$ anomaly segments with strength $\alpha$
\item Compute realized coverage $c_{\text{realized}}$
\item If $|c_{\text{realized}} - c_{\text{target}}| < \delta$: return corrupted sequence
\item Else if $c_{\text{realized}} < c_{\text{target}}$: $\alpha \gets \alpha \times U(1.1, 1.4)$
\item Else: $\alpha \gets \alpha \times U(0.7, 0.9)$
\end{enumerate}
\item Return best attempt (closest to target)
\end{enumerate}

This adaptive mechanism maintains mean realized coverage 15.2\% $\pm$ 3.1\% across batches regardless of sequence characteristics.

\section{Synthetic Test Set: Injection Formulations}
\label{app:test_injector}

This appendix provides mathematical formulations for representative anomaly types in the synthetic benchmark. Each type implements 3--4 equation variants to discourage superficial pattern matching. Additional qualitative examples, parameter tables, and reproducibility notes are available in the HydroGEM Hugging Face repository: \url{https://huggingface.co/Ejokhan/HydroGEM}.

\subsection{Drift Anomaly (4 variants)}

Drift represents gradual sensor calibration decay, documented by Shaughnessy et al.~\cite{shaughnessy2019driftr} and observed at 4.26\% prevalence in labeled USGS data~\cite{santos2024unsupervised}. Four functional forms produce visually similar monotonic departures through different mechanisms.

\textbf{Variant 1 -- Linear}:
\begin{equation}
Q'(t) = Q(t) + \beta_Q \cdot (t - t_{\text{start}}), \quad H'(t) = H(t) + \beta_H \cdot (t - t_{\text{start}})
\end{equation}
where $\beta_Q \sim U(-0.5, +0.5)$ ft$^3$/s/hr, $\beta_H \sim U(-0.01, +0.01)$ ft/hr

\textbf{Variant 2 -- Exponential}:
\begin{equation}
Q'(t) = Q(t) \cdot \exp[\alpha_Q \cdot (t - t_{\text{start}})], \quad H'(t) = H(t) \cdot \exp[\alpha_H \cdot (t - t_{\text{start}})]
\end{equation}
where $\alpha_Q \sim U(-0.01, +0.01)$ hr$^{-1}$, $\alpha_H \sim U(-0.005, +0.005)$ hr$^{-1}$

\textbf{Variant 3 -- Sigmoid}:
\begin{equation}
Q'(t) = Q(t) + \Delta Q \cdot \frac{1}{1 + \exp[-k_Q(t - t_{\text{mid}})]}, \quad H'(t) = H(t) + \Delta H \cdot \frac{1}{1 + \exp[-k_H(t - t_{\text{mid}})]}
\end{equation}
where $t_{\text{mid}} = (t_{\text{start}} + t_{\text{end}})/2$, $\Delta Q \sim U(-Q_{\text{mean}}/2, +Q_{\text{mean}}/2)$, $k_Q \sim U(0.1, 0.5)$

\textbf{Variant 4 -- Polynomial}:
\begin{equation}
Q'(t) = Q(t) + a_Q(t - t_{\text{start}})^2 + b_Q(t - t_{\text{start}}), \quad \text{similarly for } H
\end{equation}
where coefficients $a, b$ are chosen to achieve target endpoint deviation $\sim U(0.1Q_{\text{mean}}, 0.3Q_{\text{mean}})$

\subsection{Ice Backwater (3 variants)}

Ice backwater occurs when ice cover elevates stage at a given discharge, a phenomenon documented in TWRI 3-A10~\cite{kennedy1984discharge} and WSP 2175~\cite{rantz1982measurement}. The characteristic signature is elevated stage with suppressed or unchanged discharge.

\textbf{Variant 1 -- Gradual onset}:
\begin{align}
\alpha_{\text{ice}}(t) &= \alpha_{\max} \cdot \frac{t - t_{\text{start}}}{t_{\text{peak}} - t_{\text{start}}} \quad \text{for } t \in [t_{\text{start}}, t_{\text{peak}}] \\
H'(t) &= H(t) \cdot [1 + \alpha_{\text{ice}}(t)] \\
Q'(t) &= Q(t) \cdot [1 - \beta_{\text{ice}}(t)]
\end{align}
where $\alpha_{\max} \sim U(0.15, 0.55)$, $\beta_{\max} \sim U(0, 0.10)$, $t_{\text{peak}} - t_{\text{start}} \sim U(12, 48)$ hours

\textbf{Variant 2 -- Abrupt onset with gradual recovery}:
\begin{equation}
\alpha_{\text{ice}}(t) = \begin{cases}
\alpha_{\max} & t \in [t_{\text{start}}, t_{\text{recover}}] \\
\alpha_{\max} \cdot \exp[-k(t - t_{\text{recover}})] & t > t_{\text{recover}}
\end{cases}
\end{equation}
where $k \sim U(0.01, 0.05)$ hr$^{-1}$

\textbf{Variant 3 -- Periodic breakup events}:
\begin{equation}
\alpha_{\text{ice}}(t) = \alpha_{\text{base}} + \sum_{i=1}^{N_{\text{events}}} A_i \exp\left[-\frac{(t - t_i)^2}{2\sigma_i^2}\right]
\end{equation}
where $\alpha_{\text{base}} \sim U(0.2, 0.4)$, $N_{\text{events}} \sim \{2, 3, 4, 5\}$, event amplitudes $A_i \sim U(-0.3, -0.1)$ (negative for breakup), widths $\sigma_i \sim U(1, 6)$ hours

\subsection{Rating Shift (3 variants)}

Rating shift occurs when stream morphology changes abruptly, producing persistent bias in the stage-discharge relationship. Such shifts are documented in TWRI 3-A10~\cite{kennedy1984discharge}, and Mansanarez et al.~\cite{mansanarez2019shift} provide Bayesian methods for adjusting ratings to morphological changes at known times.

\textbf{Variant 1 -- Instantaneous}:
\begin{equation}
Q'(t) = Q(t) \cdot (1 + \delta), \quad t \geq t_{\text{shift}}
\end{equation}
where $\delta \sim U(0.15, 0.55) \times \text{strength}$

\textbf{Variant 2 -- Transition period}:
\begin{equation}
Q'(t) = Q(t) \cdot \left(1 + \delta \cdot \frac{t - t_{\text{shift}}}{t_{\text{settle}} - t_{\text{shift}}}\right), \quad t \in [t_{\text{shift}}, t_{\text{settle}}]
\end{equation}
where transition duration $t_{\text{settle}} - t_{\text{shift}} \sim U(6, 24)$ hours

\textbf{Variant 3 -- Partial recovery}:
\begin{equation}
Q'(t) = Q(t) \cdot (1 + \delta \cdot r(t))
\end{equation}
where $r(t)$ decays from 1.0 to $r_{\text{final}} \sim U(0.3, 0.7)$ over the affected window

\subsection{Spike (4 variants)}

Spikes are brief anomalous excursions documented at 0.13\% (large) and 0.16\% (small) prevalence in operational data~\cite{santos2024unsupervised}. Leigh et al.~\cite{leigh2019framework} include spikes in their anomaly taxonomy for water quality sensors.

\textbf{Variant 1 -- Electronic} (instantaneous):
\begin{equation}
Q'(t_{\text{spike}}) = Q(t_{\text{spike}}) + s \cdot \sigma_Q, \quad s \sim U(3, 5)
\end{equation}

\textbf{Variant 2 -- Hydraulic} (brief duration):
\begin{equation}
Q'(t) = Q(t) + A \cdot \exp\left[-\frac{(t - t_{\text{peak}})^2}{2\tau^2}\right], \quad \tau \sim U(1, 3) \text{ hours}
\end{equation}

\textbf{Variant 3 -- Additive offset}:
\begin{equation}
Q'(t) = Q(t) + \Delta, \quad \Delta \sim U(0.2, 0.5) \times Q_{\text{segment mean}}
\end{equation}

\textbf{Variant 4 -- Bounded} (capped at physical limits):
\begin{equation}
Q'(t) = \min(Q(t) + s \cdot \sigma_Q, Q_{\text{max,site}})
\end{equation}

\section{Design Rationale and Parameter Selection}
\label{app:design_rationale}

This appendix documents the rationale for key design decisions in SynthStream, grounded in USGS operational documentation and peer-reviewed literature.

\subsection{Duration Bounds}

Duration bounds for each anomaly type derive from USGS field documentation describing the temporal characteristics of sensor and hydraulic phenomena.

\begin{table}[htbp]
\centering
\caption{Duration bounds with literature basis}
\label{tab:duration_bounds}
\footnotesize
\begin{tabular}{@{}p{2.5cm}p{2.5cm}p{6.5cm}@{}}
\toprule
\textbf{Anomaly Type} & \textbf{Duration Range} & \textbf{Basis} \\
\midrule
Dropout & 1--120 hr & Telemetry outages vary from brief interruptions to multiday failures~\cite{sauer2002standards} \\
\addlinespace
Flatline & 2--144 hr & Sensor freeze events persist until maintenance intervention~\cite{sauer2010stage} \\
\addlinespace
Ice backwater & 72--520 hr & Ice cover persists days to weeks in northern systems~\cite{ashton1986ice} \\
\addlinespace
Drift & 96--400 hr & Calibration decay develops over days to weeks~\cite{shaughnessy2019driftr} \\
\addlinespace
Rating shift & 12--288 hr & Morphological changes from flood events~\cite{kennedy1984discharge,mansanarez2019shift} \\
\addlinespace
Debris effect & 2--60 hr & Temporary obstructions clear within hours to days~\cite{rantz1982measurement} \\
\bottomrule
\end{tabular}
\end{table}

\subsection{Coverage Distribution}

The coverage distribution (percentage of timesteps affected by anomalies) follows a trimodal design with a forbidden zone between 10\% and 32\%. This ensures unambiguous separation between difficulty tiers:

\begin{itemize}
\item \textbf{Light tier (3--9\%)}: Sparse anomalies testing detection sensitivity
\item \textbf{Moderate tier (32--44\%)}: Substantial anomaly presence
\item \textbf{Heavy tier (44--60\%)}: Dense anomalies testing robustness
\end{itemize}

The 10--32\% gap prevents ambiguous sequences that could fall into either category, enabling clean stratification of benchmark results by difficulty level.

\subsection{Physical Space Injection}

All anomalies are injected in physical units (ft\textsuperscript{3}/s for discharge, ft for stage) rather than normalized space. This design choice ensures that injected anomalies respect the physical coupling between stage and discharge documented in rating curve theory~\cite{kennedy1984discharge}. For example, backwater anomalies elevate stage while suppressing discharge, consistent with the hydraulic relationship $Q = f(H)$ being violated by downstream control effects~\cite{rantz1982measurement}.

\subsection{Equation Form Variation}

Each anomaly type implements 3--4 functional forms (e.g., linear, exponential, sigmoid, polynomial for drift) that produce similar visual signatures through different mathematical mechanisms. This design prevents models from learning superficial pattern templates and requires abstraction of the underlying anomaly concept. The approach follows established practice in time series anomaly detection benchmarks~\cite{schmidl2022gutentag,paparrizos2022tsbuad}.

\subsection{Single Type Sequences}

Approximately 30\% of sequences contain only one anomaly type, enabling unambiguous per-type evaluation. The remaining 70\% contain multiple types with 40\% compound overlap probability, reflecting realistic co-occurrence patterns. This split balances diagnostic clarity with operational realism.

\subsection{Climate Conditional Mixing}

Anomaly type probabilities are adjusted based on geographic and climatic characteristics of each station:

\begin{itemize}
\item Ice backwater probability increased for high latitude stations (AK, MT, WA, MN, ND, SD, WI, MI)
\item Backwater probability increased for coastal and low elevation stations
\item Debris effect probability increased for forested watersheds
\end{itemize}

These adjustments reflect documented geographic patterns in USGS operational experience~\cite{rantz1982measurement}.

\section{Canadian Data Quality Filtering}
\label{app:canadian_qc}

This appendix details quality filtering applied to Canadian ECCC stations (Section~\ref{sec:canadian}).

\subsubsection{Hydrologically-Motivated Checks}

All checks apply to corrected series only, ensuring evaluation on windows where hydrologist-edited records behave as plausible physical time series.

\paragraph{1. Sufficient variability}

\begin{equation}
\text{CV}(H_{\text{corr}}) = \frac{\sigma(H_{\text{corr}})}{\mu(H_{\text{corr}})} > 0.10
\end{equation}

\textbf{Rationale}: Removes nearly flat records where water level hardly changes (e.g., failed sensors reporting constant values, tidal gauges at slack water). Anomaly detection is ill-posed on essentially constant series.

\paragraph{2. Monotonic rating relation}

\begin{equation}
\rho_{\text{Spearman}}(H_{\text{corr}}, Q_{\text{corr}}) > 0.5
\end{equation}

\textbf{Rationale}: Enforces physically plausible behavior where larger stages generally correspond to larger discharges. Violations suggest either: (a) severe data quality issues making the window unsuitable for evaluation, or (b) complex hydraulics (backwater, tidal influence) where the station may not be appropriate for standard QA/QC methods.

\paragraph{3. Reasonable rating curve exponent}

Fit $\log Q = \log a + b \log H$ using robust regression (RANSAC), require:

\begin{equation}
b \in [0.5, 10]
\end{equation}

\textbf{Rationale}: Natural open-channel flow typically exhibits $b \in [1.5, 3.0]$ based on hydraulic geometry~\cite{chow1988applied}. We relax to [0.5, 10] to accommodate weirs ($b \approx 1.5$), rectangular channels ($b \approx 1.0$), and complex cross-sections, while excluding physically implausible extremes.

\paragraph{4. Moderate rating curve fit}

\begin{equation}
R^2 \geq 0.3
\end{equation}

\textbf{Rationale}: Deliberately modest threshold allows hydraulic complexity (hysteresis, backwater) while removing windows where stage explains essentially none of discharge variability.

\paragraph{5. Sufficient valid data fraction}

\begin{equation}
\frac{\#\{t : Q_{\text{corr},t} > 0 \text{ and } H_{\text{corr},t} > 0\}}{\#\{\text{all timesteps}\}} \geq 0.70
\end{equation}

\textbf{Rationale}: Ensures enough usable information for robust anomaly detection features. Windows with $>30\%$ missing/invalid data lack statistical power.

\paragraph{6. Limited flatline segments}

Compute consecutive differences: $\Delta H_t = |H_{\text{corr},t} - H_{\text{corr},t-1}|$. Require:

\begin{equation}
\frac{\#\{t : \Delta H_t < 0.001 \text{ ft}\}}{\#\{\text{all timesteps}\}} < 0.30
\end{equation}

\textbf{Rationale}: Natural hydrographs exhibit continuous variation. If $>30\%$ of consecutive values are identical (within 0.001 ft precision), the gauge is likely frozen or malfunctioning. This preserves low-variability baseflow periods while removing structural sensor failures.

\subsubsection{Station-Level Filtering}

After temporal alignment of the four series (stage raw/corrected, discharge raw/corrected), we compute missing data fraction:

\begin{equation}
f_{\text{missing}} = \frac{\#\{\text{timesteps missing any of 4 series}\}}{\#\{\text{all timesteps}\}}
\end{equation}

Stations with $f_{\text{missing}} > 0.05$ (losing $>5\%$ of hourly timesteps) are excluded entirely from the Canadian test set.

\textbf{Rationale}: High missing data rates indicate either: (a) inconsistent data collection between raw and corrected archives, or (b) stations undergoing major operational changes. Both scenarios compromise evaluation validity.

\subsubsection{Filtering Statistics}

Across the full ECCC archive:

\begin{itemize}
\item Initial stations with raw + corrected data: 1,847
\item After station-level filtering ($<5\%$ missing): 1,203 (65.1\% retention)
\item After window-level filtering (6 criteria): 487 stations with eligible windows (26.4\% overall retention)
\item Random sampling for evaluation: 100 stations
\end{itemize}

The conservative filtering ensures evaluation on high-quality, hydrologically meaningful windows where model performance reflects genuine anomaly detection capability rather than artifacts of poor data quality.

\section{ECCC Evaluation Metrics}
\label{app:eccc_metrics}

This appendix provides detailed evaluation methodology and supplementary results for the zero-shot Canadian station assessment (Section~\ref{subsec:canadian_results}).

\subsection{Evaluation Metrics}

Standard pointwise F1 treats each timestamp as an independent classification decision, which penalizes detections that correctly identify anomaly events but with minor temporal offset. For weakly labeled data derived from operational corrections, this limitation is particularly problematic~\cite{kim2022rigorous}. We adopt complementary metrics grounded in the time series anomaly detection literature.

\paragraph{Segment F1.} Operates at the interval level, treating contiguous anomaly timestamps as discrete events rather than independent samples~\cite{hundman2018detecting}. Segment recall measures the fraction of ground-truth intervals detected with any overlap:
\begin{equation}
\text{Recall}_{\text{segment}} = \frac{|\{G_i : \exists P_j, \text{overlap}(G_i, P_j) > 0\}|}{|G|}
\end{equation}
where $G = \{G_1, G_2, \ldots, G_n\}$ represents ground-truth intervals and $P = \{P_1, P_2, \ldots, P_m\}$ represents predicted intervals.

\paragraph{Tolerant F1.} Introduces a temporal buffer to accommodate labeling imprecision~\cite{sorbo2024navigating}. Both ground-truth and prediction masks are dilated by $\pm\tau$ hours before computing pointwise metrics:
\begin{equation}
\tilde{G} = \text{dilate}(G, \tau), \quad \tilde{P} = \text{dilate}(P, \tau)
\end{equation}

\paragraph{Weighted F1.} Weights each timestamp's contribution by correction magnitude, prioritizing operationally significant anomalies:
\begin{equation}
w_t = \max\left(\frac{|Q_{\text{cor},t} - Q_{\text{raw},t}|}{|Q_{\text{raw},t}| + \epsilon}, \frac{|H_{\text{cor},t} - H_{\text{raw},t}|}{|H_{\text{raw},t}| + \epsilon}\right)
\end{equation}

\paragraph{Range Score.} Measures coverage of ground-truth interval extents~\cite{tatbul2018precision}:
\begin{equation}
\text{Range} = \frac{1}{|G|} \sum_{G_i \in G} \frac{|G_i \cap P|}{|G_i|}
\end{equation}

\subsection{Tolerance Sensitivity Analysis}

Table~\ref{tab:tolerance_sensitivity} presents performance across buffer sizes from $\pm$1 hour to $\pm$24 hours.

\begin{table}[htbp]
\centering
\caption{Tolerant F1 performance across different buffer sizes.}
\label{tab:tolerance_sensitivity}
\small
\begin{tabular}{@{}lccc@{}}
\toprule
Buffer & Precision & Recall & F1 \\
\midrule
$\pm$1h & 0.654 & 0.590 & 0.596 \\
$\pm$2h & 0.657 & 0.607 & 0.607 \\
$\pm$4h & 0.661 & 0.634 & 0.623 \\
$\pm$6h & 0.665 & 0.654 & 0.636 \\
$\pm$12h & 0.672 & 0.702 & 0.664 \\
$\pm$24h & 0.683 & 0.765 & 0.700 \\
\bottomrule
\end{tabular}
\end{table}

Precision remains stable across all tolerance values (0.654--0.683), indicating that HydroGEM's predictions are consistently located near genuine anomalies regardless of how strictly temporal alignment is measured. Recall increases with larger tolerance (0.590 at $\pm$1h to 0.765 at $\pm$24h), reflecting that many detections fall within a day of recorded correction boundaries but not within the exact hour.

\subsection{Extended Multi-Metric Results}

Table~\ref{tab:extended_metrics} presents performance across all evaluation metrics.

\begin{table}[htbp]
\centering
\caption{Zero-shot HydroGEM performance on 100 ECCC stations across extended metrics.}
\label{tab:extended_metrics}
\small
\begin{tabular}{@{}lccc@{}}
\toprule
Metric & Precision & Recall & F1/Score \\
\midrule
Pointwise F1 & 0.650 & 0.567 & 0.582 \\
Segment F1 & 0.583 & 0.901 & 0.676 \\
Tolerant F1 ($\pm$24h) & 0.683 & 0.765 & 0.700 \\
Weighted F1 & 0.991 & 0.593 & 0.720 \\
Range Score & --- & --- & 0.647 \\
\bottomrule
\end{tabular}
\end{table}

Weighted precision of 0.991 indicates that when HydroGEM flags a severe anomaly, it is correct 99\% of the time. The Range Score of 0.647 shows that HydroGEM covers approximately 65\% of the temporal extent of anomaly intervals on average.

\section{Baseline Methods and Parameters}
\label{app:baselines}

This appendix specifies the 11 baseline detectors used in Section~\ref{subsec:baselines}. All baselines are evaluated in a strict zero-shot setting and require no labeled anomalies. Following the defendable baseline philosophy encoded in our implementation, all decision thresholds are either standard literature defaults (e.g., 3$\sigma$, Tukey fences) or computed from statistics of the evaluated sequence itself. No baseline parameters were tuned using USGS synthetic test labels or Canadian correction-derived weak labels.

\subsection{Inputs and Shared Handling}
\label{app:baseline_inputs}

All baselines operate on the corrupted hourly discharge and stage values $(Q_t, H_t)$ for each evaluated 576-hour window. Missing values are handled as follows. For multivariate baselines (Isolation Forest and LOF), missing values in each variable are imputed by the variable median within the window. For STL residual detection, the series is forward filled then backward filled; remaining missing values are filled by the within-window median.

Unless otherwise stated, decisions are computed independently for discharge and stage and combined by a logical OR.

\subsection{Statistical Baselines (3)}
\label{app:baseline_statistical}

\paragraph{Z-Score.}
For each variable independently, we compute the absolute z-score and flag an anomaly if either variable exceeds a fixed threshold:
\begin{equation}
|\text{z}(Q_t)| > 3 \;\; \text{or} \;\; |\text{z}(H_t)| > 3.
\end{equation}
The 3$\sigma$ threshold is the standard literature rule.

\paragraph{IQR (Tukey fence).}
For each variable independently, we compute the 25th and 75th percentiles $(Q_1, Q_3)$ and $\text{IQR} = Q_3 - Q_1$, then flag:
\begin{equation}
x_t < Q_1 - 1.5\,\text{IQR} \;\; \text{or} \;\; x_t > Q_3 + 1.5\,\text{IQR},
\end{equation}
with the discharge and stage masks combined by OR. If $\text{IQR}$ is near zero, no anomalies are flagged.

\paragraph{Moving average residual.}
For each variable, we compute a centered rolling mean and rolling standard deviation using a 168-hour window. Let $m_t$ be the rolling mean and $\sigma_t$ be the rolling standard deviation. We flag:
\begin{equation}
|x_t - m_t| > 3\,\sigma_t,
\end{equation}
with discharge and stage combined by OR. The 168-hour window captures weekly periodicity and the threshold is local and data derived via $\sigma_t$.

\subsection{Generic Unsupervised Baselines (3)}
\label{app:baseline_unsupervised}

\paragraph{Isolation Forest.}
We fit an Isolation Forest to the 2D feature vector $[Q_t, H_t]$ within each evaluated window after standardization using \texttt{StandardScaler}. We use 100 trees and \texttt{random\_state=42}. We set \texttt{contamination='auto'} and use the native decision function threshold at 0, flagging:
\begin{equation}
\text{decision\_function}(t) < 0.
\end{equation}
This yields an algorithm-native outlier decision without specifying an anomaly fraction.

\paragraph{Local Outlier Factor (LOF).}
We fit LOF on the standardized 2D features $[Q_t, H_t]$ within each window using $k = \min(20, n-1)$ neighbors. We compute LOF scores as $s_t = -\texttt{negative\_outlier\_factor\_}$ and flag anomalies using a data-derived threshold equal to the 95th percentile of scores within the same window:
\begin{equation}
s_t > \text{percentile}_{95}(s).
\end{equation}
This flags the top 5\% most anomalous points per window.

\paragraph{STL residual.}
For each variable independently, we apply STL decomposition with period 168 hours and \texttt{robust=True} when the series length is at least twice the period. Let $r_t$ denote the STL residual and $\sigma_r$ its standard deviation within the window. We flag:
\begin{equation}
|r_t| > 3\,\sigma_r.
\end{equation}
If the window is too short for STL, we fall back to z-score thresholding with the same 3$\sigma$ rule. Discharge and stage detections are combined by OR.

\subsection{Hydrology-Motivated Baselines (5)}
\label{app:baseline_hydrology}

\paragraph{Rating curve residual.}
Within each window, for valid points with $Q>0$ and $H>0$, we fit a power law rating relationship using log-linear regression:
\begin{equation}
Q \approx a (H - H_0)^b.
\end{equation}
We set $H_0$ using a low-stage offset estimated as the 1st percentile of $H$ minus a small constant, and fit $(a,b)$ via linear regression in log space. Let $e_t$ be the absolute log residual between observed $Q_t$ and predicted $\hat{Q}_t$. We compute the standard deviation of the fitted residuals $\sigma_e$ and flag:
\begin{equation}
e_t > 3\,\sigma_e.
\end{equation}
If insufficient valid points are available, no anomalies are flagged.

\paragraph{Rate of change.}
We compute the absolute relative change for discharge and stage:
\begin{equation}
r_t = \frac{|x_t - x_{t-1}|}{x_{t-1}},
\end{equation}
for timesteps where $x_{t-1} > 0$ and both values are defined. For each variable separately, we set a data-derived threshold as the 99th percentile of $r_t$ within the same window and flag points exceeding this threshold. Discharge and stage flags are combined by OR.

\paragraph{Persistence (stuck sensor).}
For each variable, we compute a centered rolling standard deviation using a 12-hour window with \texttt{min\_periods=win/2}. We estimate a sensor-resolution threshold as 0.1\% of the within-window data range defined by the 99th minus 1st percentile:
\begin{equation}
\tau = 0.001 \left(P_{99}(x) - P_{1}(x)\right).
\end{equation}
We flag a timestep as anomalous if the rolling standard deviation is below $\tau$. Discharge and stage are combined by OR.

\paragraph{Q--H consistency.}
We compute the centered rolling Pearson correlation between discharge and stage using a 24-hour window. We flag an anomaly when the rolling correlation is negative and below a physics-motivated threshold:
\begin{equation}
\rho_{t,24h}(Q,H) < -0.3.
\end{equation}

\paragraph{Seasonal envelope.}
We define approximate monthly bins using only the time index of the window by mapping hours to day of year and grouping into 13 coarse period bins (approximately monthly). For each bin and variable, we compute lower and upper bounds as the 1st and 99th percentiles within that bin. A timestep is flagged if it falls outside its bin-specific bounds. Discharge and stage flags are combined by OR. If a bin contains insufficient samples, it is skipped.

\subsection{Implementation Notes and Reproducibility}
\label{app:baseline_notes}

All baselines are applied to corrupted inputs to match the operational detection setting. Thresholds are fixed by literature conventions (3$\sigma$, 1.5$\times$IQR) or computed from statistics of each evaluated window (rolling standard deviation, percentile thresholds, fitted residual variance). Stochastic components (Isolation Forest) use a fixed random seed. The full reference implementation is provided in our released evaluation scripts.


\FloatBarrier  
\bibliographystyle{elsarticle-num} 
\bibliography{references}

\end{document}